\documentclass[10pt,twocolumn,letterpaper]{article}

\usepackage{wacv}              

\usepackage{graphicx}
\usepackage{amsmath}
\usepackage{amssymb}
\usepackage{booktabs}
\usepackage{amsmath,amsfonts,bm}

\newcommand{\norm}[1]{\left\lVert#1\right\rVert}

\def\Tabref#1{Table~\ref{#1}}

\def\Figref#1{Figure~\ref{#1}}

\def\eqref#1{equation~\ref{#1}}
\def\Eqref#1{Equation~\ref{#1}}

\def\ceil#1{\lceil #1 \rceil}

\def\1{\bm{1}}

\DeclareMathAlphabet{\mathsfit}{\encodingdefault}{\sfdefault}{m}{sl}
\SetMathAlphabet{\mathsfit}{bold}{\encodingdefault}{\sfdefault}{bx}{n}

\let\citep\cite

\usepackage{comment}
\usepackage{enumitem}
\usepackage{mathtools}
\usepackage{amsmath}
\usepackage{amssymb}
\usepackage{booktabs}
\usepackage{adjustbox}
\usepackage{mathtools}
\usepackage{graphicx}
\usepackage{bbm}
\usepackage{multirow}
\usepackage{capt-of,etoolbox}
\usepackage{color, colortbl}
\definecolor{LightCyan}{rgb}{0.88,1,1}
\definecolor{LightRed}{rgb}{1.0, 0.91, 0.91}
\definecolor{LightGray}{rgb}{0.88,0.88,0.88}
\definecolor{VeryLightGray}{rgb}{0.93,0.93,0.93}

\definecolor{cvprblue}{rgb}{0.21,0.49,0.74}
\usepackage[pagebackref,breaklinks,colorlinks,citecolor=cvprblue]{hyperref}

\usepackage[capitalize]{cleveref}
\crefname{section}{Sec.}{Secs.}
\Crefname{section}{Section}{Sections}
\Crefname{table}{Table}{Tables}
\crefname{table}{Tab.}{Tabs.}

\def\wacvPaperID{2243} 
\def\confName{WACV}
\def\confYear{2025}

\begin{document}

\title{Improving Shift Invariance in Convolutional Neural Networks with \\ Translation Invariant Polyphase Sampling}

\author{
Sourajit Saha\\
University of Maryland, Baltimore County\\
{\tt\small ssaha2@umbc.edu}
\and 
Tejas Gokhale\\
University of Maryland, Baltimore County\\
{\tt\small gokhale@umbc.edu}
}

\maketitle

\begin{abstract}
Convolutional neural networks (CNNs), widely deployed in several applications, contain downsampling operators in their pooling layers which have been observed to be sensitive to pixel-level shift, affecting the robustness of CNNs.
We study shift invariance through the lens of \textit{maximum-sampling bias (MSB)} and find MSB to be negatively correlated with shift invariance.
Based on this insight, we propose a learnable pooling operator called \textit{Translation Invariant Polyphase Sampling (TIPS)} to reduce MSB and learn translation-invariant representations.
TIPS results in consistent performance gains on multiple benchmarks for image classification, object detection, and semantic segmentation in terms of accuracy, shift consistency, shift fidelity, as well as improvements in adversarial and distributional robustness.
TIPS results in the lowest MSB compared to all previous methods, thus explaining the strong empirical results. 
TIPS can be integrated into any CNN and can be trained end-to-end with marginal computational overhead.
Code: \href{https://github.com/sourajitcs/tips/}{https://github.com/sourajitcs/tips/}
\end{abstract}

\section{Introduction} \label{sec:intro}

Shift invariance is an ideal property for visual recognition models to ensure that outputs remain invariant to small pixel-level shifts in input images. 
Shift-invariance is desirable for image classification to ensure that categorical outputs are not affected by small horizontal and/or vertical pixel shift, and shift-equivariance is desirable for object detection and semantic segmentation to ensure that pixel-shift in the image results in equivalent shift in outputs. 
Recent studies have also found shift-invariant vision models to be more robust on \textit{out-of-distribution} testing and adversarial attacks, making shift invariance an important facet of trustworthiness and risk assessment for visual recognition.

Although convolution layers are shift-equivariant~\cite{lecun1989handwritten}, recent studies~\cite{azulay2019deep, gao2019lip} reveal that pooling operators such as max pooling, average pooling, and strided convolution break shift invariance by violating the Nyquist sampling theorem~\cite{nyquist1928certain}.
Since these pooling operators sample the highest feature activation in a pooling window, we study if this design choice can explain model's performance under pixel shift, by introducing the concept of Maximum-Sampling Bias (MSB) which denotes the tendency to pool larger values within pooling windows. 
We observe a strong negative correlation between MSB and shift invariance, i.e.\ models with higher MSB are the least shift invariant.

\begin{figure}[t]
    \centering
    \includegraphics[width=\linewidth,trim={0 0em 0 0},clip]{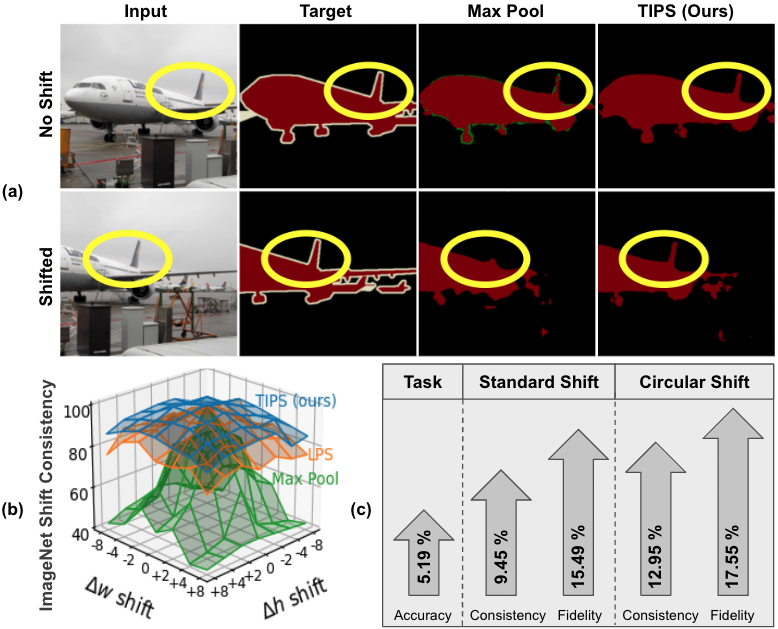}
    \caption{
    Translation-Invariant Polyphase Sampling (TIPS) is a pooling operator that improves shift invariance of CNNs.
    \textbf{(a)} An illustration of the improvements in semantic segmentation with TIPS;
    \textbf{(b)} Greater shift consistency of TIPS at higher degrees of pixel shift;
    \textbf{(c)} TIPS results in consistent and architecture-agnostic improvements in accuracy and four measures of shift invariance for image classification and semantic segmentation.
    }
    \label{fig:tease}
\end{figure}

\begin{figure*}[t]
    \centering
    \includegraphics[width=0.92\linewidth]{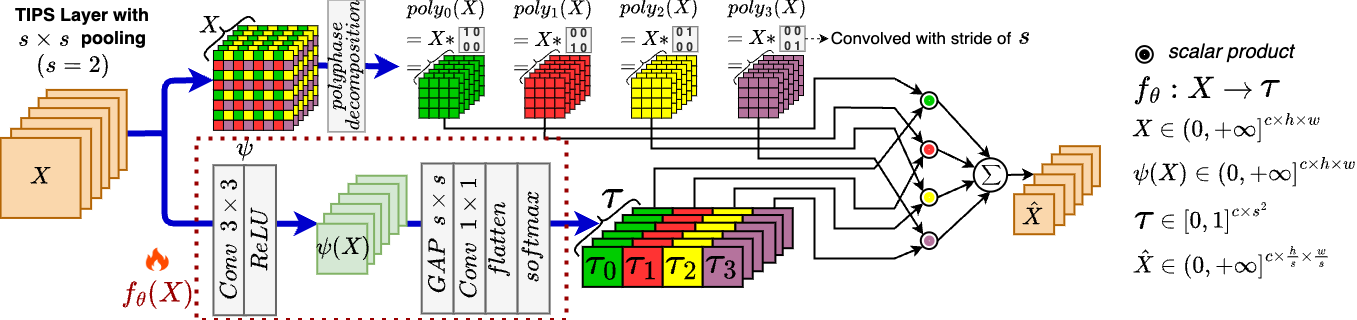}
    \caption{
        TIPS downsamples ReLU-activated intermediate feature map $X$ into $\hat{X}$ with stride $s$ and 
        learns polyphase mixing coefficients $\tau$ (using a small fully convolutional function $f_{\theta}$)
        which results in the output feature map as the weighted linear combination $\hat{X}$.
        The polyphase decomposition on input feature map $X$ results in $\mathrm{poly}\strut_{i}$ which are then mixed as a weighted linear combination with $\tau$ to compute $\hat{X}$.
    }
    \label{fig:tips}
\end{figure*}

Informed by this finding, we design a pooling operator, Translation Invariant Polyphase Sampling (TIPS), to discourage MSB and improve invariance under shift transformations. 
We introduce two loss functions: $\mathcal{L}_{FM}$ -- to discourage known \textit{failure modes} of shift invariance and $\mathcal{L}_{undo}$ -- to learn to \textit{undo} standard shift.
We demonstrate that this approach consistently improves robustness under shift transformation on multiple image classification, object detection, and semantic segmentation benchmarks, outperforming data augmentation and contrastive learning strategies, and resulting in state-of-the-art performance in terms of accuracy, shift consistency, and shift fidelity under standard and circular shift transformations, while operating at a small computational overhead.
In real world scenarios, standard shifts are more likely to occur than circular shifts; however, current literature largely focuses on circular shift invariance.
We show that our method improves shift invariance on both circular shift and standard shift (shown in \Figref{fig:tease}). 
When tested on adversarial attacks, patch attacks, and natural corruptions, models trained with TIPS exhibit greater robustness than previous pooling operators.

\section{Related Work}  \label{sec:lit}

\noindent\textbf{Robustness of CNNs} has been examined under geometric transformations~\citep{cohen2016group,engstrom2019exploring}, domain shift~\citep{venkateswara2017deep}, attribute shift~\citep{gokhale2021attribute}, adversarial attacks~\citep{madry2018towards}, and natural corruptions~\citep{hendrycks2018benchmarking}.
Distributional robustness of CNNs has been explored through random or learned data augmentation~\citep{xu2020robust,gokhale2023improving}, contrastive learning~\citep{khosla2020supervised}, and Bayesian approaches~\citep{cheng2023adversarial}.

\smallskip\noindent \textbf{Dense Sampling and Anti-aliasing.} Conventional sliding window downsampling in computer vision algorithms is typically applied with stride that is bigger than $1$ which breaks shift equivariance~\citep{simoncelli1992shiftable}. 
Shift invariance can be improved through dense sampling~\citep{leung2001representing} with dilated convolutions~\citep{yu2017dilated} with susceptibility to griding artifacts. 
Zhang \etal\cite{zhang2019making} suggest BlurPool to enhance shift invariance through anti-aliasing before downsampling, whereas Zou \etal\cite{zou2020delving} propose DDAC, to learn low pass anti-aliasing filter. 

\smallskip\noindent \textbf{Polyphase Sampling.} 
APS~\citep{chaman2021truly} and LPS~\citep{rojas2022learnable} use polyphase sampling to satisfy the Nyquist sampling theorem~\citep{nyquist1928certain} and permutation invariance which provides robustness against circular shifts. 
APS enhances shift invariance by sampling the highest energy polyphase index ($\ell_{p}$ norm) while LPS learns the sampling. 
LPS being sensitive to gumble softmax temperature can sample polyphases to maximize downstream objective which does not consider shift invariance unless training data is shift-augmented. 
Evidence ~\cite{zhang2019making, zou2020delving, chaman2021truly, rojas2023making} suggests that although polyphase sampling methods can improve shift invariance for circular shift, they still struggle to deal with standard shift; we show that TIPS can benefit both.
The focus of this study is to improve robustness of CNNs against both standard and circular shifts, which constitute significant aspects of model evaluation.

\begin{figure}[t]
    \centering
    \includegraphics[width=\linewidth,trim={0 0em 0 0},clip]{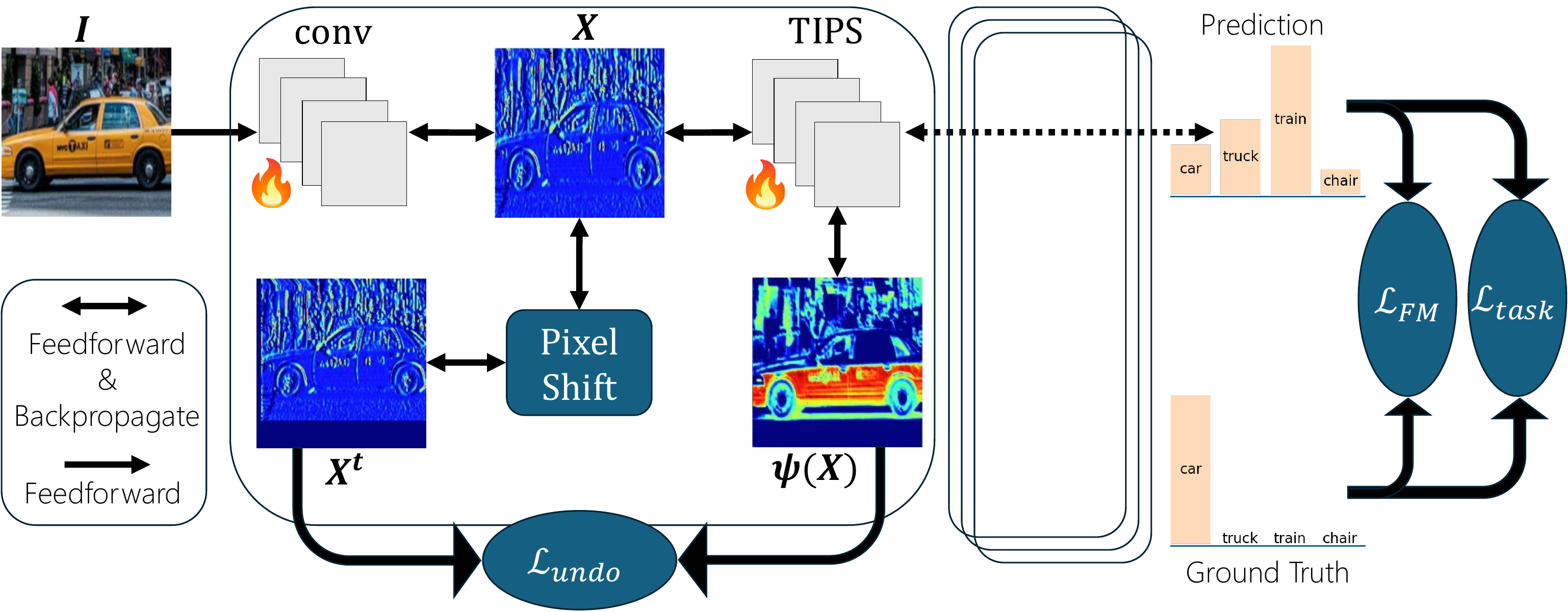}
    \caption{The \textit{end-to-end} training pipeline with TIPS, regularization to undo shift $\mathcal{L}_{undo}$, regularization to discourage known failure modes $\mathcal{L}_{FM}$, and downstream task loss $\mathcal{L}_{task}$.
    }
    \label{fig:ti_loss_exp}
\end{figure}

\begin{figure*}[t]
    \centering
    \includegraphics[width=\linewidth,trim={0 6pt 0 0},clip]{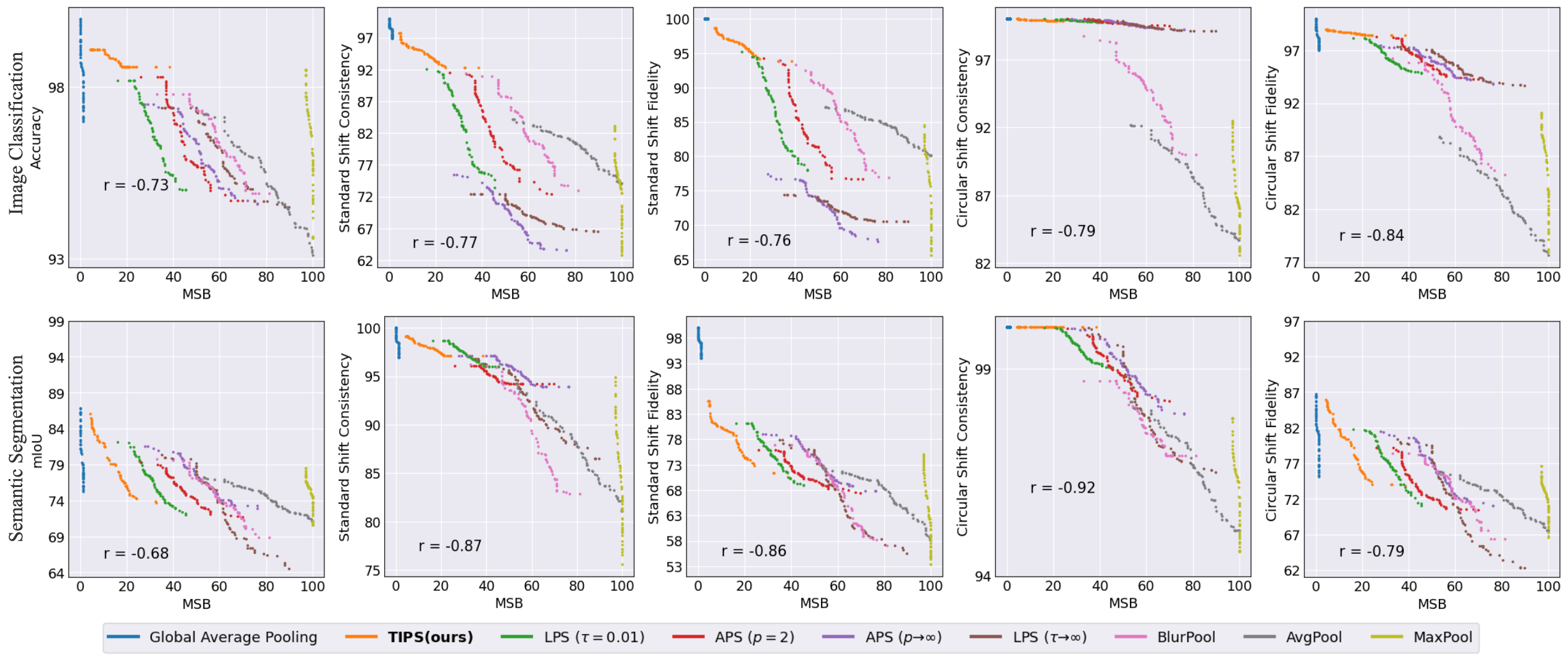} 
    \caption{
        Our large-scale correlation study 
        shows a strong negative correlation of performance with MSB (\%) as indicated by Pearson's $r$. 
        Linear clusters with negative correlation are also observed for points belonging to each pooling method. 
    }
    \label{fig:msb_cons}
\end{figure*}

\section{Translation Invariant Polyphase Sampling} \label{sec:method}
In this section, we discuss the design of our proposed TIPS pooling layer and the workflow to train CNNs with TIPS.
Let $X{\in}\mathbb{R}^{c\times h\times w}$ be a ReLU-activated feature map, where $c,~h,~w$ are the number of channels, height, and width.
A pooling layer with sampling rate or stride $s$ (also called $s$$\times$$s$ pooling) downsamples $X$ into $\hat{X}$.

\subsection{TIPS: A Learnable Pooling Layer} \label{sec:tips}
TIPS \textit{learns} to sample polyphase decompositions of input feature maps $X$ using two branches, as shown in \Figref{fig:tips}.
In the first branch, polyphase components of $X$ with stride $s$ are computed similar to Chaman \etal\cite{chaman2021truly}:
{\small
\begin{align}
    & \mathrm{poly}_{is+j}(X) = X[k, s n_1 + i, s n_2 + j], \\
    & \forall~ 
    i, j{\in}\mathbb{Z}\strut^{s-1}_{0}; k{\in}\mathbb{Z}\strut^{c-1}_{0};
    n_1{\in}\mathbb{Z}^{\ceil{\frac{h}{s}}}_{0}; n_2{\in}\mathbb{Z}^{\ceil{\frac{w}{s}}}_{0}. \nonumber
\end{align}
}
Note that polyphase sampling can also be achieved by a strided convolution with a $s\times s$ kernel equal to 1 at index $(i, j)$ and $0$ elsewhere.
A visualization of polyphase sampling with $s=2$ is shown in \Figref{fig:tips}.
The second branch is a function $f_\theta:{X}{\rightarrow}\tau$ that learns mixing coefficients $\tau{\in}[0, 1]^{c\times s^2}$.
$\psi()$ is a $3\times3$ convolutional layer followed by ReLU activation.
All operations in $f_\theta$ are shift invariant.
The output of TIPS layer is computed as a weighted linear combination of the polyphase components of it's input:
{\small
\begin{equation}
    \hat{X} = \sum_{i, j} \tau\strut_{is+j}~\mathrm{poly}\strut_{is+j}(X)
\label{eq:linear_combo}
\end{equation}
}

\smallskip\noindent\textbf{Regularizing TIPS to Discourage Known Failure Modes}
Chaman \etal\cite{chaman2021truly} have shown that skewed mixing coefficients (\eg $\tau{=}\{0,1,0,0\}$ for $s{=}2$) are not robust against standard shift.
TIPS with uniform mixing coefficients and LPS with higher softmax temperature (\eg $\tau{=}\{0.25, 0.25, 0.25, 0.25\}$ for $s{=}2$) is identical to average pooling, which hurts shift invariance~\citep{zhang2019making}.
Based on these known failure modes, we introduce a regularization on the mixing coefficients in TIPS -- the first term discourages skewed $\tau$ and the second term discourages uniform $\tau$.
{\small
\begin{equation}
    \mathcal{L}_{FM} = 
    (\norm{\tau}_{2} - 1) + (1 - s^2\norm{\tau}_{2}) = (1-s^2)\norm{\tau}_{2}. 
    \label{eq:l_si}
\end{equation}
}

\smallskip\noindent\textbf{Learning to Undo Standard Shift} \label{sec:tiloss}
Although prior work have shown the benefits of using polyphase sampling to counter circular shift, performance still degrades with standard shifts due to information loss beyond the pooling boundary.
To address this, we shift the input feature map with a random amount of vertical and horizontal standard shift sampled from uniform distribution $U(0, \frac{h}{10})$ and $U(0, \frac{w}{10})$ respectively to obtain a shifted $X^t$ and design a regularization with the objective of undoing this shift:
{\small
\begin{equation}
\mathcal{L}_{undo} = \mathbbm{E}_{h^{'} \in h} \mathbbm{E}_{w^{'} \in w} [X^{t}_{h^{'},w^{'}} - \psi(X_{h^{'},w^{'}})]^2
\label{eq:tips_loss}
\end{equation}
}

\subsection{Training CNNs with TIPS} \label{sec:train_pipe}
Let $N$ be the number of training epochs.
For the first $\epsilon N$ epochs, we train only with the task loss $\mathcal{L}_{task}$ and the regularization to discourage failure modes $\mathcal{L}_{FM}$.
For subsequent epochs, the undo regularization $\mathcal{L}_{undo}$ is introduced.
The final training loss is the Lagrangian (with $\alpha{\in}[0, 1]$):
{\small
\begin{equation}
\mathcal{L} = 
    (1-\alpha) \mathcal{L}_{task} + \mathbbm{1}(epoch < \epsilon N) \alpha \mathcal{L}_{undo} + \mathcal{L}_{FM}
\label{eq:train_loss}
\end{equation}
}
We gradually introduce $\mathcal{L}_{undo}$ to stabilize training so that the downstream performance is not compromised. 
$\psi(X)$ contains a $3\times3$ convolution layer, ReLU, followed by global average pooling, $1\times 1 $ convolution, flattening, and a softmax operation to obtain $\tau$ as shown in \Figref{fig:tips}.
Weights $\theta$ are Kaiming-normal initialized \citep{he2016deep}.

\begin{table*}[t]
\resizebox{\linewidth}{!}{
\centering
\subfloat[TinyImageNet]{
\resizebox{0.5\linewidth}{!}{
\Large
\begin{tabular}{@{}cllllll@{}}
\toprule
& & \multicolumn{1}{c}{\textbf{Unshifted}} & \multicolumn{2}{c}{\textbf{Standard Shift}} & \multicolumn{2}{c}{\textbf{Circular Shift}} \\
\cmidrule(lr){3-3} \cmidrule(lr){4-5} \cmidrule(lr){6-7}
 &\textbf{Method} & \textbf{Acc.} $\uparrow$ & \textbf{Consistency} $\uparrow$ & \textbf{Fidelity} $\uparrow$ & \textbf{Consistency} $\uparrow$ & \textbf{Fidelity} $\uparrow$ \\
\midrule
\multirow{8}{*}{\rotatebox[origin=c]{90}{CNN (ResNet-101)}} & MaxPool & 78.54{\footnotesize$\pm$0.22} & 88.45{\footnotesize$\pm$0.15} & 69.47{\footnotesize$\pm$0.14} & 92.82{\footnotesize$\pm$.14} & 79.20{\footnotesize$\pm$0.15} \\
& APS & 83.01{\footnotesize$\pm$0.08} & 91.37{\footnotesize$\pm$0.06} & 75.85{\footnotesize$\pm$0.04} & \textbf{100.00{\footnotesize$\pm$0.00}} & 83.01{\footnotesize$\pm$0.08} \\
& LPS & 85.67{\footnotesize$\pm$0.18} & 92.95{\footnotesize$\pm$0.04} & 79.63{\footnotesize$\pm$0.05} & \textbf{100.00{\footnotesize$\pm$0.00}} & 85.67{\footnotesize$\pm$0.18} \\
& \textbf{TIPS} & \textbf{86.78{\footnotesize$\pm$0.19}} & \textbf{94.27{\footnotesize$\pm$0.14}} & \textbf{81.80{\footnotesize$\pm$0.15}} & \textbf{100.00{\footnotesize$\pm$0.00}} & \textbf{86.78{\footnotesize$\pm$0.19}} \\ 
\cmidrule(lr){2-7}
& BlurPool (LPF-5) & 82.83{\footnotesize$\pm$0.13} & 90.81{\footnotesize$\pm$0.17} & 75.22{\footnotesize$\pm$0.12} & 95.87{\footnotesize$\pm$0.19} & 79.41{\footnotesize$\pm$1.12} \\
& APS (LPF-5) & 83.52{\footnotesize$\pm$0.03} & 92.00{\footnotesize$\pm$0.20} & 76.84{\footnotesize$\pm$0.11} & \textbf{100.00{\footnotesize$\pm$0.00}} & 83.52{\footnotesize$\pm$0.03} \\ 
& LPS (LPF-5) & 86.74{\footnotesize$\pm$0.09} & 93.38{\footnotesize$\pm$0.17} & 80.99{\footnotesize$\pm$0.04} & \textbf{100.00{\footnotesize$\pm$0.00}} & 86.74{\footnotesize$\pm$0.09} \\
& \textbf{TIPS (LPF-5)} & \textbf{86.91{\footnotesize$\pm$0.13}} & \textbf{94.55{\footnotesize$\pm$0.06}} & \textbf{88.20{\footnotesize$\pm$0.07}} & \textbf{100.00{\footnotesize$\pm$0.00}} & \textbf{86.91{\footnotesize$\pm$0.13}} \\
\midrule
\multirow{2}{*}{\rotatebox[origin=c]{90}{ViT}} & ViT-B/16 (I21k) & 89.34{\footnotesize$\pm$0.06} & 73.47{\footnotesize$\pm$0.03} & 65.64{\footnotesize$\pm$0.11} & 72.94{\footnotesize$\pm$0.19} & 65.16{\footnotesize$\pm$0.08} \\ 
& ViT-L/16 (I21k) & 90.75{\footnotesize$\pm$0.15} & \textbf{74.39{\footnotesize$\pm$0.16}} & \textbf{67.51{\footnotesize$\pm$0.14}} & 73.85{\footnotesize$\pm$0.06} & 67.02{\footnotesize$\pm$0.19} \\ 
& Swin-B (I21k) & \textbf{91.19{\footnotesize$\pm$0.04}} & 72.14{\footnotesize$\pm$0.15} & 65.78{\footnotesize$\pm$0.17} & \textbf{75.49{\footnotesize$\pm$0.05}} & \textbf{68.84{\footnotesize$\pm$0.10}} \\ 
\bottomrule
\end{tabular}
}
}
\subfloat[ImageNet]{
\resizebox{0.5\linewidth}{!}{
\Large
\begin{tabular}{@{}cllllll@{}}
\toprule
& & \multicolumn{1}{c}{\textbf{Unshifted}} & \multicolumn{2}{c}{\textbf{Standard Shift}} & \multicolumn{2}{c}{\textbf{Circular Shift}} \\
\cmidrule(lr){3-3} \cmidrule(lr){4-5} \cmidrule(lr){6-7}
 &\textbf{Method} & \textbf{Acc.} $\uparrow$ & \textbf{Consistency} $\uparrow$ & \textbf{Fidelity} $\uparrow$ & \textbf{Consistency} $\uparrow$ & \textbf{Fidelity} $\uparrow$ \\
\midrule
\multirow{8}{*}{\rotatebox[origin=c]{90}{CNN (ResNet-101)}} & MaxPool & 76.31{\footnotesize$\pm$0.18} & 89.05{\footnotesize$\pm$0.19} & 67.05{\footnotesize$\pm$0.06} & 87.56{\footnotesize$\pm$0.13} & 66.82{\footnotesize$\pm$0.17} \\
& APS & 76.07{\footnotesize$\pm$0.15} & 90.95{\footnotesize$\pm$0.13} & 69.19{\footnotesize$\pm$0.13} & \textbf{100.00{\footnotesize$\pm$0.00}} & 76.07{\footnotesize$\pm$0.15} \\
& LPS & 78.29{\footnotesize$\pm$0.14} & 91.74{\footnotesize$\pm$0.03} & 71.82{\footnotesize$\pm$0.13} & \textbf{100.00{\footnotesize$\pm$0.00}} & 78.29{\footnotesize$\pm$0.14} \\
& \textbf{TIPS} & \textbf{80.24{\footnotesize$\pm$0.09}} & \textbf{92.87{\footnotesize$\pm$0.08}} & \textbf{74.52{\footnotesize$\pm$0.18}} & \textbf{100.00{\footnotesize$\pm$0.00}} & \textbf{80.24{\footnotesize$\pm$0.09}} \\   
\cmidrule(lr){2-7}
& BlurPool (LPF-5) & 76.33{\footnotesize$\pm$0.08} & 90.70{\footnotesize$\pm$0.14} & 69.23{\footnotesize$\pm$0.15} & 90.55{\footnotesize$\pm$0.17} & 69.12{\footnotesize$\pm$0.19} \\
& APS (LPF-5) & 76.49{\footnotesize$\pm$0.08} & 91.23{\footnotesize$\pm$0.17} & 69.78{\footnotesize$\pm$0.05} & 99.98{\footnotesize$\pm$0.00} & 76.41{\footnotesize$\pm$0.06} \\
& LPS (LPF-5) & 78.31{\footnotesize$\pm$0.05} & 92.49{\footnotesize$\pm$0.15} & 72.43{\footnotesize$\pm$0.08} & \textbf{100.00{\footnotesize$\pm$0.00}} & 78.31{\footnotesize$\pm$0.05} \\  
& \textbf{TIPS (LPF-5)} & \textbf{81.36{\footnotesize$\pm$0.10}} & \textbf{93.11{\footnotesize$\pm$0.03}} & \textbf{75.75{\footnotesize$\pm$0.14}} & \textbf{100.00{\footnotesize$\pm$0.00}} & \textbf{81.36{\footnotesize$\pm$0.10}} \\ 
\midrule
\multirow{2}{*}{\rotatebox[origin=c]{90}{ViT}} & ViT-B/16 (I21k) & 83.89{\footnotesize$\pm$0.07} & 84.38{\footnotesize$\pm$0.05} & 70.79{\footnotesize$\pm$0.27} & 81.03{\footnotesize$\pm$0.11} & 67.98{\footnotesize$\pm$0.19} \\ 
& ViT-L/16 (I21k) & 85.06{\footnotesize$\pm$0.02} & 83.19{\footnotesize$\pm$0.12} & 70.76{\footnotesize$\pm$0.17} & 81.64{\footnotesize$\pm$0.15} & 69.44{\footnotesize$\pm$0.14} \\ 
& Swin-B (I21k) & \textbf{85.16{\footnotesize$\pm$0.05}} & \textbf{85.24{\footnotesize$\pm$0.19}} & \textbf{72.59{\footnotesize$\pm$0.05}} & \textbf{82.79{\footnotesize$\pm$0.08}} & \textbf{70.50{\footnotesize$\pm$0.18}} \\ 
\bottomrule
\end{tabular}
}
}
}
\caption{Image classification performance on TinyImageNet and ImageNet averaged over five trials.}
\label{tab:cls_tinyimgnet}
\end{table*}

\section{Maximum-Sampling Bias}
In this section, we set up a framework to study shift invariance in CNNs, by defining maximum-sampling bias (MSB).
Through a large-scale analysis, we show that MSB is negatively correlated with shift invariance.

\smallskip\noindent\textbf{Definition of MSB} \label{sec:msb}
Existing pooling operators exhibit a common tendency to propagate signals based on activation strength.
We denote this phenomenon as maximum-sampling bias (MSB), defined as the fraction of window locations for which the maximum signal value is sampled;
a higher MSB indicates a higher probability of maximum signal values being selected during pooling. 

Let $p()$ be a pooling operator in a CNN and $s$ be the downsampling factor; \eg $s{=}2$ for max-pooling with a $2{\times}2$ window. 
Let $X{\in}\mathbb{R}^{h \times w}$ be the input to a pooling layer. 
Applying a pooling operator $p()$ on $X$ with downsampling factor $s$ results in an output $\hat{X} = p(X){\in}\mathbb{R}^{\frac{h}{s} \times \frac{w}{s}}$. 
It is trivial to see that $MSB=1$ for max-pool, as max pool selects $\hat{X}[i, j]{=}{\mathrm{max}_{m,n}}~X[is{+}m, js{+}n]$.
Average pooling produces $X[i, j]{=}{\mathbb{E}_{m,n}} x[i s + m, j s + n]$ and is equivalent to max-pooling if all values within the window are identical. 
For all other cases, the average value (less than the maximum) is sampled
and thus $MSB \leq 1$.
APS pooling~\citep{chaman2021truly} samples the polyphase component of $X$ with maximum $\ell_{p}$ norm; $\hat{X}[i, j]{=}{\mathrm{max}_{is+j}}\, \{ \lVert poly_{j}(x) \rVert_{p} \}_{i,j=0}^{s-1}$.
As the polyphase function $poly()$ in APS and LPS is a monotonic function, it also exhibits a preference for sampling larger signals in the pooling window.

\smallskip\noindent\textbf{Negative Correlation between MSB and Shift Invariance.} \label{sec:corr}
We investigate whether MSB is linked to shift invariance, by conducting a large-scale analysis of correlation between MSB and shift invariance on a number of visual recognition benchmarks with multiple CNN architectures and pooling methods
\footnote{see supp.mat. for details on experiment setup for correlation analysis.}.
We evaluated 576 models across different architectures, datasets, and pooling methods
and conducted a correlation study as shown in \Figref{fig:msb_cons} with MSB on x-axis and performance metrics on the y-axis for both image classification and semantic segmentation. 
We observe a strong negative correlation between MSB and shift invariance metrics and also with downstream task performance (accuracy, mIoU).
Circular consistency is more negatively correlated with MSB than standard consistency. 
Linear relationships are observed for points corresponding to each pooling method across architectures and datasets.
Using Global Average Pooling (GAP)~\citep{he2016deep} with no spatial downsampling of the intermediate feature maps helps shift invariance, but with significant computational costs.
TIPS achieves high shift invariance with marginal computational overhead (see supplementary material for comparison).

\section{Experiments} \label{sec:exp}
We perform experiments on multiple benchmarks for both image classification, object detection, and semantic segmentation.
For image classification, we evaluate shift invariance while for object detection and semantic segmentation we evaluate shift equivariance. 
However, following conventions used in prior work~\cite{zou2020delving, rojas2022learnable}, this is also referred to as ``shift invariance'' in the semantic segmentation results.
For both classification, detection, and segmentation, we avoid using pre-trained CNNs since the pre-training step use non-shift invariant strided convolution and maxpool.

\subsection{Image Classification Experiments} \label{sec:cls}
\noindent\textbf{Datasets and Baselines} 
We benchmark the performance of TIPS and prior work on five image classification datasets:
CIFAR-10 ~\cite{krizhevsky2009learning}, Food-101~\cite{bossard2014food}, Oxford-102~\cite{nilsback2008automated}, Tiny ImageNet~\cite{le2015tiny}, and ImageNet~\cite{krizhevsky2012imagenet}. 
Results for Food-101 and Oxford-102 are in the Supplementary Material.
Our baselines include MaxPool, APS (\begin{small}$p$$=$$2$\end{small}), and LPS (\begin{small}$\tau$$=$$0.23$\end{small}), as well as BlurPool, APS, and LPS with anti-aliasing using $n$$\times$$n$ Gaussian low-pass filter (LPF-5).
We also compare with three Vision Transformer (ViT) architectures: ViT-B/16, ViT-L-16~\cite{dosovitskiy2020image}, and Swin-B~\cite{liu2021swin} which are pre-trained on the larger ImageNet-21k dataset~\cite{deng2009imagenet}.

\smallskip\noindent\textbf{Hyperparameters} 
For TIPS, we choose $\epsilon = 0.4$ and $\alpha=0.35$ in \Eqref{eq:train_loss}.
All models are trained using an SGD optimizer with initial learning rate 0.05, momentum 0.9, and weight decay 1e-4 with early stopping. 
No models in our experiments were trained on shifted images.
For each dataset-backbone pair, for fair comparison, TIPS and all baselines are trained with identical hyperparameters.

\smallskip\noindent\textbf{Evaluation Metrics} 
In addition to reporting accuracy on the unshifted test set, we use the \textit{consistency} definition from~\cite{zou2020delving} which  compares the predictions for two shifted images.
However, as \textit{consistency} does not consider the ground truth label ($y$) for evaluation, we introduce \textit{fidelity} as a new metric. 
Note: $x^{h_1, w_1}$ denotes image $x$ shifted by $h \sim U(0, \frac{h}{8})$ vertically and $w \sim U(0, \frac{w}{8})$ horizontally.
{\small
\begin{align}
    & \mathrm{Consistency} = { \underset{x}{\mathbb{E}}~~\underset{\substack{(h_1, w_1),(h_2, w_2)}}{\mathbb{E}} } { \mathbbm{1} } {  [f(x^{h_1, w_1})
    = f(x^{h_2, w_2}) ] \nonumber } \\
    & \mathrm{Fidelity} = {  \underset{x}{\mathbb{E}}~~\underset{\substack{\scriptstyle(h_1, w_1),(h_2, w_2)}}{\mathbb{E}} } { \displaystyle \mathbbm{1} } { [y = f(x^{h_1, w_1}) 
    = f(x^{h_2, w_2}) ] \nonumber }    
\end{align}
}

\smallskip\noindent\textbf{Results} 
Tables~\ref{tab:cls_tinyimgnet} and \ref{tab:cls_cifar_c10} show strong dataset- and backbone-agnostic evidence for the efficacy of TIPS in terms of accuracy and shift invariance for both standard shift and circular shift.
TIPS results in large gains in consistency and fidelity on standard shift, which was a challenge for prior work.
It is important to note that TIPS with LPF-5 also improves upon prior work that uses LPF-5 anti-aliasing.
For ViTs, shift invariance performance is inferior to CNNS, even though they consistently achieve higher accuracy.
ViT architectures - despite being pre-trained on a very large scale dataset ImageNet21k (I21k) does not improve shift invariance which indicates that large-scale pre-training has no implications on shift invariance. 
While CNNs in general perform better on circular shift than standard shift, there is no such clear trend for ViT -- for example, ViTs are more robust on standard shift for Oxford-102 and Tiny ImageNet and more robust on circular shift for the remaining datasets.

\begin{table}[t]
    \centering
    \Large
    \resizebox{\linewidth}{!}{
    \begin{tabular}{@{}cllllll@{}}
        \toprule
        & & \multicolumn{1}{c}{\textbf{Unshifted}} & \multicolumn{2}{c}{\textbf{Standard Shift}} & \multicolumn{2}{c}{\textbf{Circular Shift}} \\
        \cmidrule(lr){3-3} \cmidrule(lr){4-5} \cmidrule(lr){6-7}
         &\textbf{Method} & \textbf{Acc.} $\uparrow$ & \textbf{Consistency} $\uparrow$ & \textbf{Fidelity} $\uparrow$ & \textbf{Consistency} $\uparrow$ & \textbf{Fidelity} $\uparrow$ \\
        \midrule
        \multirow{8}{*}{\rotatebox[origin=c]{90}{CNN (ResNet-18)}} 
        & MaxPool & 91.43{\footnotesize$\pm$0.04} & 87.43{\footnotesize$\pm$0.05} & 79.94{\footnotesize$\pm$0.05} & 90.18{\footnotesize$\pm$0.03} & 82.45{\footnotesize$\pm$0.08} \\
        & APS & 94.02{\footnotesize$\pm$0.07} & 92.89{\footnotesize$\pm$0.08} & 87.33{\footnotesize$\pm$0.05} & \textbf{100.00{\footnotesize$\pm$0.00}} & 94.02{\footnotesize$\pm$0.07} \\
        & LPS & 94.45{\footnotesize$\pm$0.05} & 93.11{\footnotesize$\pm$0.07} & 87.94{\footnotesize$\pm$0.03} & \textbf{100.00{\footnotesize$\pm$0.00}} & 94.45{\footnotesize$\pm$0.05} \\
        & \textbf{TIPS} & \textbf{95.75{\footnotesize$\pm$0.11}} & \textbf{98.38{\footnotesize$\pm$0.37}} & \textbf{94.20{\footnotesize$\pm$0.08}} & \textbf{100.00{\footnotesize$\pm$0.00}} & \textbf{95.75{\footnotesize$\pm$0.11}} \\
        \cmidrule(lr){2-7}
        & BlurPool (LPF-5) & 94.29{\footnotesize$\pm$0.11} & 91.04{\footnotesize$\pm$0.09} & 85.84{\footnotesize$\pm$0.12} & 98.27{\footnotesize$\pm$0.11} & 92.66{\footnotesize$\pm$0.07} \\
        & APS (LPF-5) & 94.44{\footnotesize$\pm$0.09} & 93.25{\footnotesize$\pm$0.13} & 88.06{\footnotesize$\pm$0.17} & \textbf{100.00{\footnotesize$\pm$0.00}} & 94.44{\footnotesize$\pm$0.09} \\
        & LPS (LPF-5) & 95.17{\footnotesize$\pm$0.12} & 94.87{\footnotesize$\pm$0.08} & 90.09{\footnotesize$\pm$0.15} & \textbf{100.00{\footnotesize$\pm$0.00}} & 95.17{\footnotesize$\pm$0.12} \\
        & \textbf{TIPS (LPF-5)} & \textbf{96.05{\footnotesize$\pm$0.13}} & \textbf{98.65{\footnotesize$\pm$0.11}} & \textbf{94.75{\footnotesize$\pm$0.10}} & \textbf{100.00{\footnotesize$\pm$0.00}} & \textbf{96.05{\footnotesize$\pm$0.13}} \\
        \midrule
        \multirow{3}{*}{\rotatebox[origin=c]{90}{ViT}} & ViT-B/16 (I21k) & 98.89{\footnotesize$\pm$0.04} & 82.34{\footnotesize$\pm$0.07} & 81.43{\footnotesize$\pm$0.05} & 83.79{\footnotesize$\pm$0.15} & 82.86{\footnotesize$\pm$0.12} \\
         & ViT-L/16 (I21k) & 99.15{\footnotesize$\pm$0.02} & 82.72{\footnotesize$\pm$0.09} & 82.01{\footnotesize$\pm$0.08} & \textbf{84.41{\footnotesize$\pm$0.11}} & \textbf{83.69{\footnotesize$\pm$0.06}} \\
         & Swin-B (I21k) & \textbf{{99.22{\footnotesize$\pm$0.03}}} & \textbf{{83.19{\footnotesize$\pm$0.07}}} & \textbf{{82.54{\footnotesize$\pm$0.05}}} & 84.05{\footnotesize$\pm$0.04} & 83.40{\footnotesize$\pm$0.04} \\
        \bottomrule
    \end{tabular}
    }
    \caption{Image classification performance on CIFAR-10.
    }
    \label{tab:cls_cifar_c10}
\end{table}

\smallskip\noindent\textbf{Comparison with other strategies for improving shift invariance.} \label{sec:other}
Besides pooling, we compare with methods that use data augmentation while training: standard shift, circular shift, and their combination, and two contrastive learning approaches: SimCLR~\citep{chen2020simple} (self-supervised) and SupCon~\citep{khosla2020supervised} (supervised) with the objective of aligning shifted samples closer and are used for downstream image classification.
\Tabref{tab:other_methods_cls} shows that 
while data augmentation does not significantly improve performance, both contrastive learning methods outperform MaxPool on all metrics.
TIPS, without any contrastive learning or data augmentation, results in greater shift invariance.

\smallskip\noindent\textbf{Applicability of TIPS to Vision Transformers} \label{sec:lim}
We focus on improving shift invariance of CNNs -- models that are widely used in real-world applications.  
Our method is based on polyphase downsampling and cannot be directly extended to vision transformers (ViT).
However, in Tables \ref{tab:cls_tinyimgnet} and \ref{tab:cls_cifar_c10}, we demonstrate that ViTs are not shift invariant and our simple plug-in solution for CNNs (TIPS) achieves higher shift invariance than ViTs.
We note that in vision transformers, three modules break shift invariance: 
\begin{itemize}[leftmargin=*,nosep,noitemsep]
    \item \textit{Patch embeddings} convert image patches into vectors using strided convolution (not shift invariant).
    \item \textit{Positional encodings} for both shifted and non-shifted inputs are identical (amount of shift is not encoded).
    \item \textit{Window-based self-attention} is computationally cheap, but applying local attention on window sizes larger than the amount of input shift causes token values to vary.
\end{itemize}

\begin{table}[t]
    \centering
    \Huge
    \resizebox{\linewidth}{!}{
    \begin{tabular}{llccccc}
    \toprule
    \multicolumn{2}{c}{}   & \multicolumn{1}{c}{\textbf{Unshifted}} & \multicolumn{2}{c}{\textbf{Standard Shift}} & \multicolumn{2}{c}{\textbf{Circular Shift}} \\
    \cmidrule(lr){3-3} \cmidrule(lr){4-5} \cmidrule(lr){6-7}
    \textbf{Strategy}   & \textbf{Method} & \textbf{Acc.} $\uparrow$ & \textbf{Consistency} $\uparrow$ & \textbf{Fidelity} $\uparrow$ & \textbf{Consistency} $\uparrow$ & \textbf{Fidelity} $\uparrow$ \\
    \midrule
    \textit{Pooling}                               & MaxPool & 64.88 & 82.41 & 53.14 &  80.39 & 50.71 \\
    \textit{(no anti-}            
                                                   & DDAC & 67.59 & 85.43 & 57.74 &  80.90 & 54.68 \\ 
     \textit{aliasing)}                           & APS & 67.05 & 86.39 & 57.92 & \cellcolor{LightCyan}{\textbf{100.00}} & 67.05 \\
                                                    & LPS & 67.39 & 86.17 & 58.07 & \cellcolor{LightCyan}{\textbf{100.00}} & 67.39 \\
                                          & \textbf{TIPS} & \textbf{69.02} & \textbf{87.42} & \textbf{60.34} & \cellcolor{LightCyan}{\textbf{100.00}} & \textbf{69.02} \\
    \midrule
    \textit{Pooling}                   & BlurPool & 66.85 & 87.43 & 58.54 &  87.88 & 58.75 \\
    \textit{(with LPF-5)}              & DDAC     & 66.98 & 86.92 & 58.22 &  \cellcolor{LightRed}{80.35} & 53.82 \\ 
    \textit{(anti-aliasing)}           & APS      & 67.52 & 87.02 & 58.76 &  99.98 & 67.51 \\
                                       & LPS      & 69.11 & 86.58 & 59.84 & \cellcolor{LightCyan}{\textbf{100.00}} & 69.11 \\
                                       & \textbf{TIPS} & \textbf{70.01} & \cellcolor{LightCyan}{\textbf{87.51}} & \textbf{61.27} & \cellcolor{LightCyan}{\textbf{100.00}} & \cellcolor{LightCyan}{\textbf{70.01}} \\
    \midrule
    \textit{Data Aug.}                  & circular & \cellcolor{LightRed}{64.25} & 83.58 & 53.71 &  84.27 & 54.14 \\
                                                & standard & \cellcolor{LightRed}{63.91} & 84.45 & 53.97 &  81.27 & 51.94 \\
                                                & both     & \textbf{64.87} & \textbf{84.99} & \textbf{55.13} &  \textbf{85.64} & \textbf{55.55} \\
    \midrule
    \textit{Contrastive}                & SimCLR & 71.15 & 85.63 & 60.93 &  \cellcolor{LightRed}{78.26} & 55.68 \\
    \textit{Learning}                   & SupCon & \cellcolor{LightCyan}{\textbf{72.49}} & \textbf{86.17} & \cellcolor{LightCyan}{\textbf{62.46}} &  \textbf{81.75} & \textbf{59.26} \\
    \bottomrule
    \end{tabular}
    }
    \caption{Comparison of ResNet18 trained on ImageNet using pooling, data augmentation, and contrastive learning.
    Best per section: \textbf{bold}; lower than MaxPool: \colorbox{LightRed}{red}; overall best: \colorbox{LightCyan}{cyan}.
    }
    \label{tab:other_methods_cls}
\end{table}

{
\begin{table*}[t]
\centering
\large
\resizebox{\linewidth}{!}{
\begin{tabular}{lcllllllllll}
\toprule
\multicolumn{1}{c}{} & \multicolumn{1}{c}{} & \multicolumn{5}{c}{\textbf{PASCAL VOC 2012 - DeepLabV3+ (ResNet-18) }} & \multicolumn{5}{c}{\textbf{Cityscapes - DeepLabV3+ (ResNet-101)}} \\
\cmidrule(lr){3-7} \cmidrule(lr){8-12}
\multicolumn{2}{c}{} & \multicolumn{1}{c}{\textbf{Unshifted}} & \multicolumn{2}{c}{\textbf{Standard Shift}} & \multicolumn{2}{c}{\textbf{Circular Shift}} & \multicolumn{1}{c}{\textbf{Unshifted}} & \multicolumn{2}{c}{\textbf{Standard Shift}} & \multicolumn{2}{c}{\textbf{Circular Shift}} \\
\cmidrule(lr){3-3} \cmidrule(lr){4-5} \cmidrule(lr){6-7} \cmidrule(lr){8-8} \cmidrule(lr){9-10} \cmidrule(lr){11-12}
\textbf{Method} & \textbf{Anti-Alias} & \textbf{mIOU} $\uparrow$ & \textbf{Consistency} $\uparrow$ & \textbf{Fidelity} $\uparrow$ & \textbf{Consistency} $\uparrow$ & \textbf{Fidelity} $\uparrow$ & \textbf{mIOU} $\uparrow$ & \textbf{Consistency} $\uparrow$ & \textbf{Fidelity} $\uparrow$ & \textbf{Consistency} $\uparrow$ & \textbf{Fidelity} $\uparrow$ \\
\midrule
MaxPool       &   -       & 70.03 & 95.17 & 66.65 & 95.42 & 66.82 & 78.50 & 96.03 & 75.38 & 97.07 & 76.20 \\
Blurpool      & LPF-3     & 71.02 & 95.52 & 67.84 & 96.03 & 68.20 & 78.90 & 96.09 & 75.82 & 97.94 & 77.27 \\
DDAC          & LPF-3     & 72.28 & 96.77 & 69.95 & 95.98 & 69.37 & 79.52 & 96.28 & 76.54 & 98.21 & 78.09 \\
APS           & LPF-3     & 72.37 & 97.05 & 70.24 & 96.70 & 69.98 & 79.84 & 97.53 & 77.87 & 98.32 & 78.50 \\
LPS           & LPF-3     & 72.37 & 97.98 & 70.92 &\textbf{100.00} & 72.37 & 80.15 & 98.60 & 79.03 &\textbf{100.00} & 80.15 \\
\textbf{TIPS} & LPF-3     & \textbf{73.84} & \textbf{98.65} & \textbf{72.84} &\textbf{100.00} & \textbf{73.84} & \textbf{81.37} & \textbf{99.02} & \textbf{80.57} &\textbf{100.00} & \textbf{81.37} \\
\bottomrule
\end{tabular}
}
\caption{
Semantic segmentation performance on Pascal VOC and Cityscapes datasets.}
\label{tab:seg1}
\end{table*}
\begin{figure*}[!ht]
    \centering
    \includegraphics[width=\linewidth,trim={37pt 21em 0 0},clip]{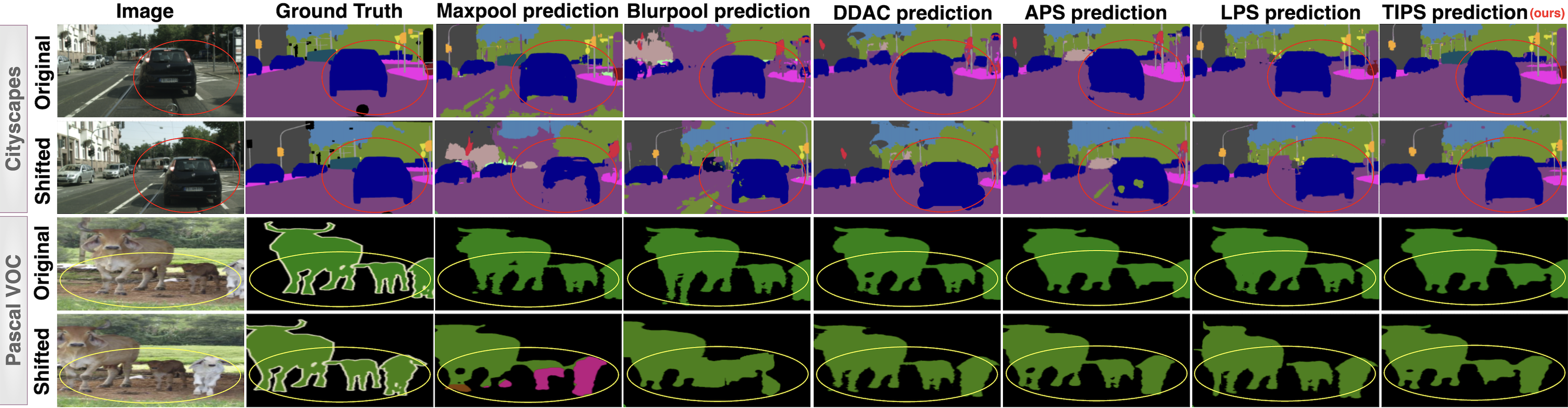}
    \caption{
    Qualitative comparison of segmentation masks predicted on original and shifted images. 
    Regions where TIPS achieve improvements (i.e. consistent segmentation quality) under linear shifts are highlighted with circles.
    }
    \label{fig:seg}
\end{figure*}
\begin{table*}[!ht]
\centering
\large
\resizebox{\linewidth}{!}{
\begin{tabular}{lllllllllllllllll}
\toprule
\multicolumn{1}{c}{} & \multicolumn{8}{c}{\textbf{RetinaNet~\cite{lin2017focal}}} & \multicolumn{8}{c}{\textbf{FasterRCNN~\cite{ren2016faster}}} \\
\cmidrule(lr){2-9} \cmidrule(lr){10-17}
\multicolumn{1}{c}{}  & \multicolumn{4}{c}{\textbf{Standard Shift}} & \multicolumn{4}{c}{\textbf{Circular Shift}} & \multicolumn{4}{c}{\textbf{Standard Shift}} & \multicolumn{4}{c}{\textbf{Circular Shift}} \\
\cmidrule(lr){2-5} \cmidrule(lr){6-9} \cmidrule(lr){10-13} \cmidrule(lr){14-17} 
\textbf{Method} & $\mathbf{AP}$ $\uparrow$ & $\mathbf{\Delta AP}$ $\downarrow$ & $\mathbf{AP_{50}}$ $\uparrow$ & $\mathbf{\Delta AP_{50}}$ $\downarrow$ & $\mathbf{AP}$ $\uparrow$ & $\mathbf{\Delta AP}$ $\downarrow$ & $\mathbf{AP_{50}}$ $\uparrow$ & $\mathbf{\Delta AP_{50}}$ $\downarrow$ & $\mathbf{AP}$ $\uparrow$ & $\mathbf{\Delta AP}$ $\downarrow$ & $\mathbf{AP_{50}}$ $\uparrow$ & $\mathbf{\Delta AP_{50}}$ $\downarrow$ & $\mathbf{AP}$ $\uparrow$ & $\mathbf{\Delta AP}$ $\downarrow$ & $\mathbf{AP_{50}}$ $\uparrow$ & $\mathbf{\Delta AP_{50}}$ $\downarrow$  \\
\midrule
MaxPool        & 36.5           & 2.2           & 56.7           & 5.1           & 36.4        & 2.3           & 56.7          & 5.4          & 37.6          & 2.9          & 59.0          & 6.4                    & 37.8           & 3.4          & 59.3              & 4.3        \\ 
BlurPool       & 35.2           & 1.4           & 55.1           & 3.4           & 36.3        & 1.6           & 56.3          & 5.4          & 37.8          & 2.0          & 58.7          & 4.6                    & 38.3           & 3.7          & 59.5          & 1.3            \\ 
APS            & 36.8           &  1.2          & 56.7           & 4.6           & 37.4        & 2.1           & 56.9          & 4.5          & 38.5          & 1.8          & 59.4          & 2.2                    & 38.3           & 3.6          & 59.6          & 1.1            \\ 
LPS            & 36.9           & 1.1           & \textbf{56.8}  & 3.8           & 37.5        & \textbf{1.9}  & 57.2          & 4.2          & 38.5          & 1.7          & 60.3          & 1.1                    & 38.7           & \textbf{2.4} & 59.9          & 1.1            \\ 
\textbf{TIPS}  & \textbf{37.0}  & \textbf{0.9}  & \textbf{56.8}  & \textbf{2.7}  & \textbf{38.0} & 2.0           & \textbf{57.4} & \textbf{3.7} & \textbf{38.6} & \textbf{1.3} & \textbf{60.5} & \textbf{1.0} & \textbf{38.9}  & 2.5          & \textbf{60.1} & \textbf{0.8}   \\ 
\bottomrule
\end{tabular}
}
\caption{
Standard and Circular Shift Equivariance for Object Detection on MS-COCO~\cite{lin2014microsoft} validation set with RetinaNet, FasterRCNN.}
\label{tab:objDet_full}
\end{table*}
}

\subsection{Semantic Segmentation Experiments} \label{sec:seg}
\noindent\textbf{Datasets and Baselines} 
We use the following datasets: PASCAL VOC 2012~\cite{everingham2010pascal}, Cityscapes~\cite{cordts2016cityscapes}, Kvasir~\cite{jha2020kvasir}, and CVC-ClinicDB~\cite{bernal2015wm}.
Results for Kvasir and CVC-ClinicDB are in the Supplementary Material.
Our baselines include MaxPool, APS, LPS, BlurPool (LPF-3), and DDAC (groups $g{=}8$, LPF-3).
BlurPool and DDAC~\cite{zou2020delving} perform antialiasing by using a fixed low-pass filter and a learnable low pass group-wise convolution filter, respectively.

\smallskip\noindent\textbf{Hyperparameters}
We use SGD optimizer with an initial learning rate 0.01, momentum 0.9, weight decay 5e-4 with early stopping. 
We use DeepLabV3+~\cite{chen2018encoder} with ResNet-18 backbone for the Pascal-VOC dataset and with ResNet-101 backbone for the Cityscapes dataset.
For Kvasir and CVC-ClinicDB, we use a UNet~\cite{ronneberger2015u} model with ``Kaiming Normal'' initialization.
For TIPS, we use $\epsilon = 0.4, \alpha = 0.35$.

\smallskip\noindent\textbf{Evaluation Metrics} 
For semantic segmentation, we compute consistency and fidelity by comparing the overlapping window for images with and without shift.
Within the common crop, we compute the percentage of pixels for which the predicted category matches the ground truth.

\smallskip\noindent\textbf{Results} 
Comparison of mIOU, consistency and fidelity in Table~\ref{tab:seg1}
shows that TIPS improves mIOU in comparison to all baselines on all four benchmarks. 
Consistent with our finding in image classification, we observe a sharper increase in shift consistency under standard shift than with circular. 
Models trained with TIPS pooling have higher fidelity on both standard and circular shifts, depicting the efficacy of TIPS in learning both shift invariant and high quality segmentation. 
In \Figref{fig:seg}, we compare the quality of the masks predicted on shifted images when using prior pooling operators and TIPS.
The areas highlighted with red circles demonstrate that TIPS segments objects with higher consistency than other pooling operators under pixel shifts. 

\subsection{Object Detection Experiments} \label{sec:obj}

\noindent\textbf{Datasets, Baselines, and Hyperparameters} 
Although there is limited work studying shift equivariance in object detection, we evaluated the efficacy of TIPS by extending the experimental settings from Manfredi \etal\cite{manfredi2020shift} to evaluate both standard and circular shift equivariance analogous to our semantic segmentation experiments. 
We train RetinaNet~\cite{lin2017focal} and FasterRCNN~\cite{ren2016faster}  on the MS-COCO dataset~\cite{lin2014microsoft} 
and present our findings in Table~\ref{tab:objDet_full}.

\medskip\noindent\textbf{Evaluation Metrics and Results.} 
In Table~\ref{tab:objDet_full} we report $AP$, 
$\Delta AP (=\textrm{best} $AP$ - \textrm{worst} $AP$)$, $AP_{50}$,
$\Delta AP_{50} (= \textrm{best} AP_{50} - \textrm{worst} AP_{50})$
for both standard and circular shift invariance. Where higher value of $AP$ and $AP_{50}$ denotes superior object detection performance and lower value of $\Delta AP$ and $ AP_{50}$ denotes superior shift equivariant object detection performance. 
Table~\ref{tab:objDet_full} demonstrates that TIPS consistently achieves lower $\Delta AP_{50}$  and higher $AP_{50}$  which demonstrates that TIPS is also effective for shift equivariant object detection in comparison to existing pooling methods.

\begin{figure*}[t]
    \centering
    \includegraphics[width=\linewidth,trim={1em 1em 1em 1em},clip]{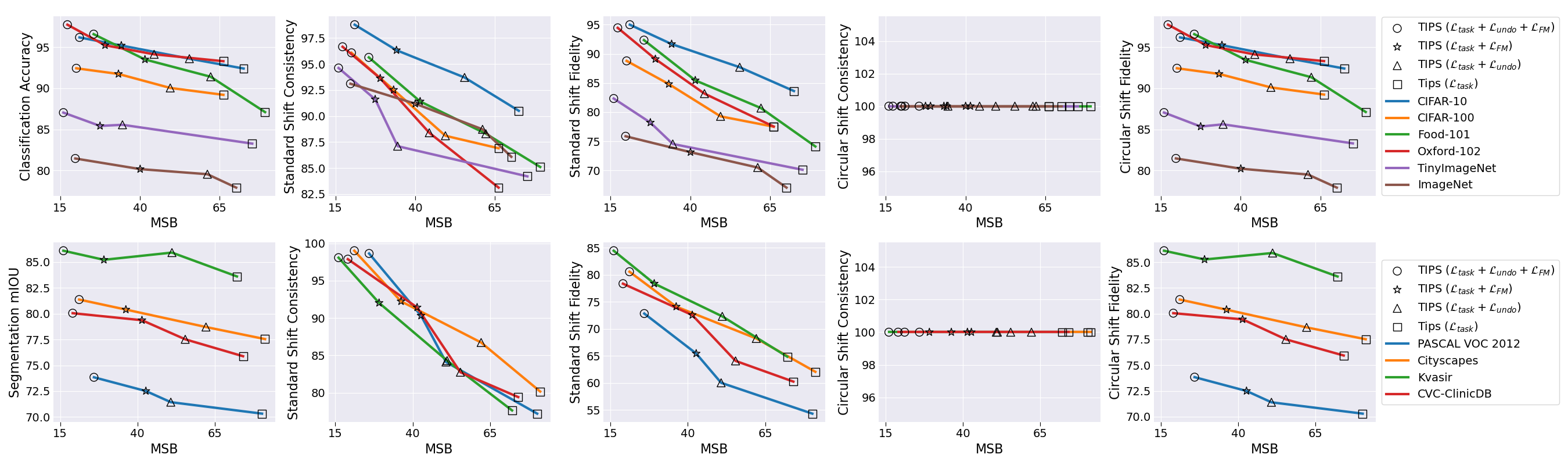}
    \caption{
    The top row shows results for six image classification datasets and the bottom row shows results for four semantic segmentation datasets.
    Regularization using both $\mathcal{L}_{undo}$ and $\mathcal{L}_{FM}$ results in the best performance.
    }
    \label{fig:abl_with_loss}
\end{figure*}

\begin{figure}
    \centering
    \includegraphics[width=1\linewidth]{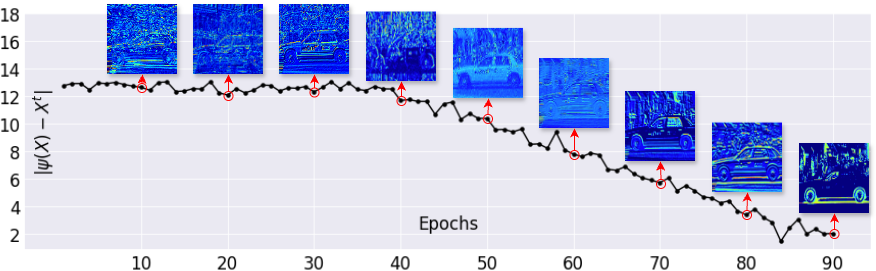}
    \caption{
    The effect of $\mathcal{L}_{undo}$ in terms of $|\psi(X)-X^t|$ and example feature maps (ResNet-101 with TIPS trained on ImageNet).
    }
    \label{fig:ti_loss_effect}
\end{figure}

\begin{figure}
    \centering
    \includegraphics[width=1\linewidth,trim={0 0em 0 0em},clip]{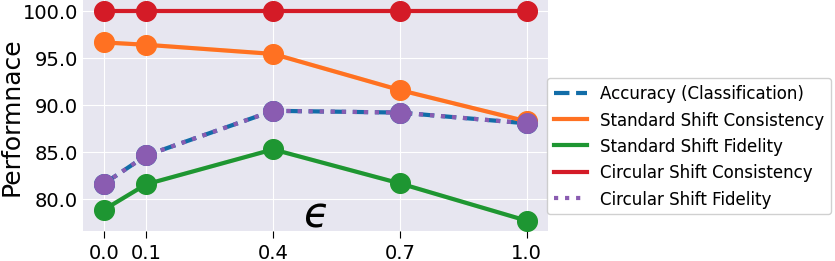}
    \caption{
    $\epsilon=0.4$ (i.e. training without $\mathcal{L}_{undo}$ for the first 40\% of epochs and with $\mathcal{L}_{undo}$ for the rest of the epochs) is optimal.
    $\epsilon{>}0.4$ yields suboptimal shift invariance, but is better than low values of $\epsilon$, demonstrating the impact of $\mathcal{L}_{undo}$.
    }
\label{fig:loss_ti}
\end{figure}

\section{Analysis}  \label{sec:ana}
We analyze the impact of our regularization through ablation studies, the impact of the choice of hyperparameters, and the effect of TIPS on various measures of robustness.

\subsection{Effect of \texorpdfstring{$\mathcal{L}_{undo}$}{L_undo} and \texorpdfstring{$\mathcal{L}_{FM}$}{L_FM} Regularization} \label{sec:effect_Lfm}

\Figref{fig:ti_loss_effect} shows the impact of $\mathcal{L}_{undo}$ by visualizing $|\psi(X)-X^t|$ at different training stages.
As training progresses, $\mathcal{L}_{undo}$ pushes $|\psi(X)-X^t|$ closer to $0$, guiding TIPS to offset standard shift on intermediate features. 
\Figref{fig:loss_ti} quantifies this observation with accuracy and shift invariance measurements with different $\epsilon$ (optimal $\epsilon{=}0.4$).

We perform an ablation study to examine the efficacy of each term in the loss function from \Eqref{eq:train_loss}.
Our results in \Figref{fig:abl_with_loss} reveal a clear trend: $\circ{>}\star{>}\triangle{>}\square$.
Across all datasets and tasks, using both regularizations together (denoted by $\circ$) leads to the highest accuracy, highest shift invariance, and lowest MSB.
Moreover, only using one of $\mathcal{L}_{undo}$ or $\mathcal{L}_{FM}$ also improves performance compared to $\mathcal{L}_{task}$.
These results demonstrate the impact of each component of our loss function on shift invariance and the inverse relationship between MSB and shift invariance.

\begin{table}
\centering
\resizebox{\linewidth}{!}{
\begin{tabular}{@{}lcccc@{}}
\toprule
              & \multicolumn{2}{c}{\textbf{MaxPool}} & \multicolumn{2}{c}{\textbf{TIPS}} \\ 
\cmidrule(lr){2-3} \cmidrule(lr){4-5}
\textbf{CNN}  & \textbf{Param.} $\downarrow$ & \textbf{Consistency} $\uparrow$ & \textbf{Param.} $\downarrow$ & \textbf{Consistency} $\uparrow$ \\ 
\midrule
ResNet-18                                  & \textbf{11.9} & 87.43          & \textbf{12.5} & \textbf{98.65}  \\ 
DenseNet-BC (k=24)~\cite{huang2017densely} & 15.3          & \textbf{90.02} & 16.9          & 96.71           \\ 
ResNet-34                                  & 21.3          & 88.93          & 22.3          & 98.60           \\ 
EfficientNet-B7~\cite{tan2019efficientnet} & 64.0          & 89.14          & 67.3          & 93.67           \\ 
\bottomrule
\end{tabular}
}
\caption{
Simply increasing the number of parameters in CNN models does not improve shift invariance. However, using TIPS as the pooling operator leads to architecture-agnostic gains.
}
\label{tab:params_extra_tab}
\end{table}

\begin{figure*}[t]
    \centering
    \includegraphics[width=1\linewidth]{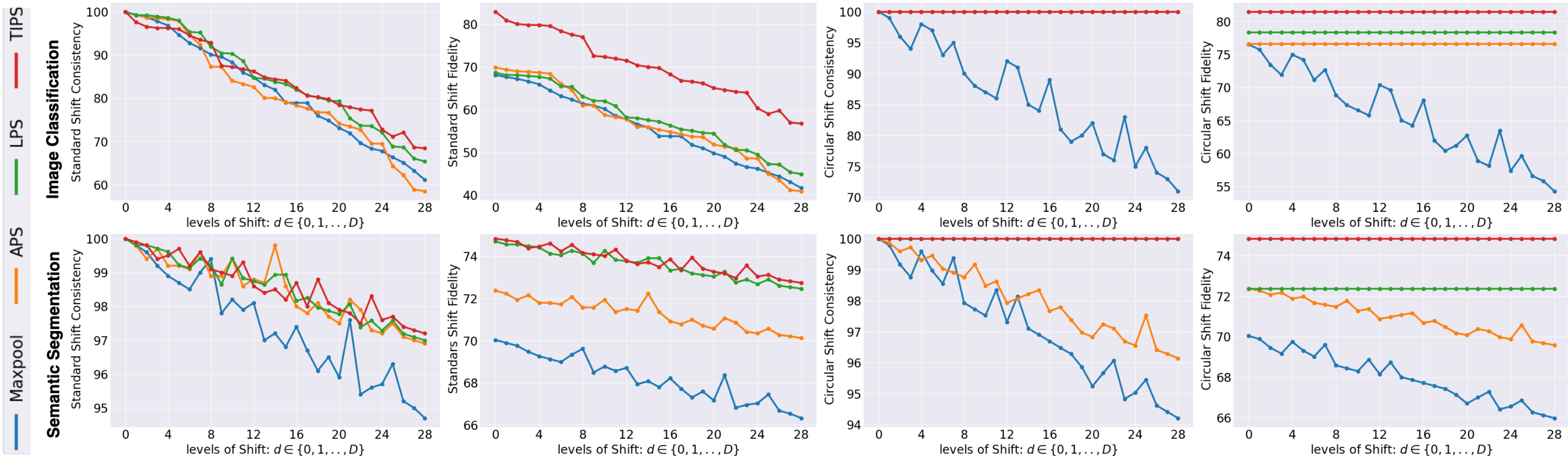}
    \caption{
    Shift invariance (consistency and fidelity) for standard and circular shifts on ImageNet classification and Cityscapes segmentation for varying degrees of shift. 
    TIPS outperform previous methods in all evaluation metrics for shift invariance.
    }
    \label{fig:area_all_exp}
\end{figure*}
\begin{figure*}[!ht]
    \centering
    \includegraphics[width=\linewidth,trim={0 0em 0 0em},clip]{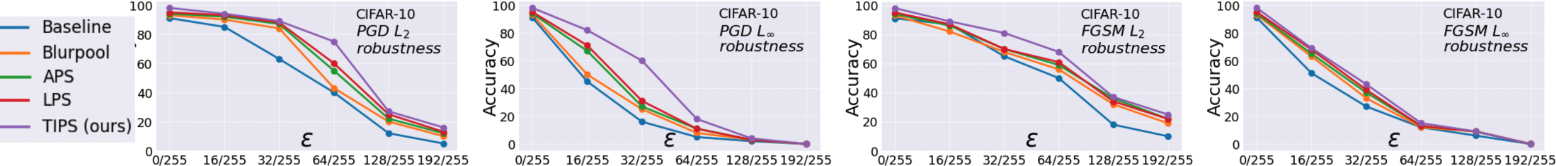}    
    \includegraphics[width=\linewidth,trim={0 12pt 0 5pt},clip]{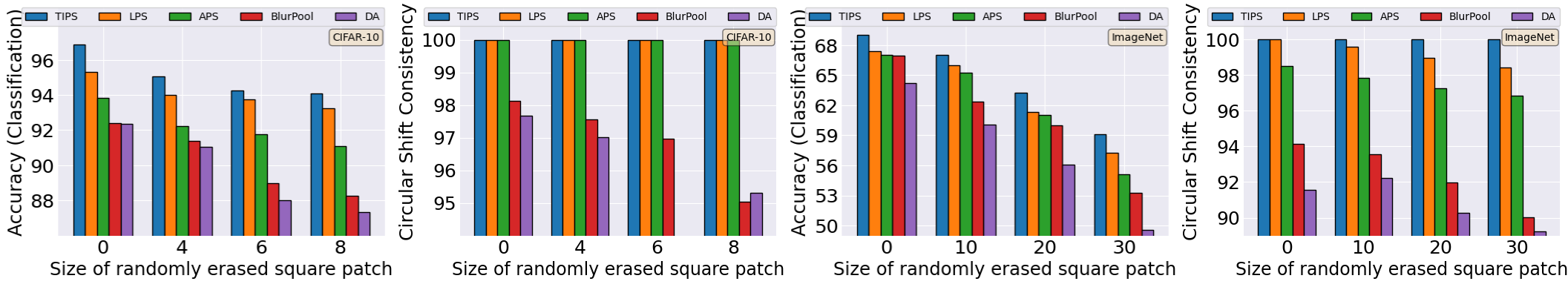}
    \caption{Evaluation of robustness of image classifiers. \textbf{Top:} Adversarial robustness under different levels (${\varepsilon}$) of input perturbations. 
    \textbf{Bottom:} Shift invariance under \textit{patch attacks} (randomly erasing image patches). 
    TIPS improves robustness of CNNs on both metrics.
    } 
    \label{fig:adv}
\end{figure*}

\subsection{Size of Models and Number of TIPS Layers}
\begin{figure}[t]
    \centering
    \includegraphics[width=1\linewidth,trim={0 0em 0 0em},clip]{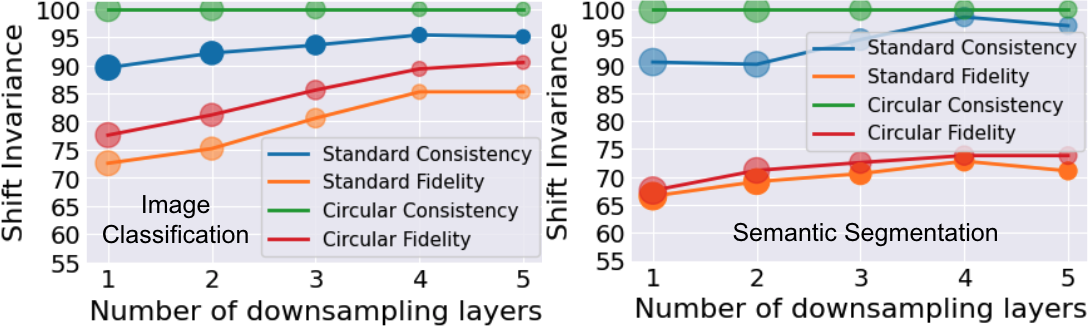}
    \caption{
        With more downsampling layers, MSB (bubble diameter in the plot) decreases and shift invariance increases.
    } 
    \label{fig:abl}
\end{figure}

Table~\ref{tab:params_extra_tab} shows that with max pooling, even a $>$ 400\% increase in parameters (from 11.9M in ResNet18 to 64.0M in EfficientNet) only results in a 1.71\% improvement in shift consistency. TIPS with ResNet18 (a 5\% increase in number of parameters from 11.9M to 12.5M), leads to a 11.22\% increase in shift consistency. 
This is observed across the board for different architectures.
\Figref{fig:abl} shows that as we train with more TIPS layers, shift invariance increases and MSB decreases, further indicating the efficacy of TIPS.

\subsection{Performance at Different Levels of Shift}
In \Figref{fig:area_all_exp}, we demonstrate shift invariance under all possible levels of shifts $d{\in}\{0,1,...,D\}$, and observe that circular shift consistency drops slower for TIPS than previous methods at higher degrees of shift.
TIPS outperforms other pooling methods on average and at all degrees of shifts in terms of shift invariance as well as downstream accuracy.

\subsection{Robustness Evaluation}   \label{sec:ood}
\noindent\textbf{Adversarial Attacks} \label{sec:adv}
Recent studies reveal the vulnerability of CNNs against adversarial attacks if they are optimized for domain generalization~\citep{frei2023double} or shift invariance~\citep{singla2021shift}, and a trade-off between adversarial robustness and other forms of generalization~\citep{gokhale2022generalized, moayeri2022explicit, teney2024id}.
We investigate $\ell_2$ and $\ell_\infty$ robustness of TIPS (with ResNet-34 backbone) using PGD and FGSM attacks from Foolbox~\citep{rauber2017foolbox}.
\Figref{fig:adv} shows TIPS to have superior adversarial robustness compared to previous methods; better shift invariance is generally correlated with better adversarial robustness.

\smallskip\noindent\textbf{Patch Attacks}
We adopt the experiment setup from~\cite{chaman2021truly} where square patches are randomly erased from the input image and test models trained on the clean CIFAR-10 and ImageNet datasets using a ResNet-18 backbone. 
\Figref{fig:adv} demonstrates that TIPS outperforms other methods (pooling and data augmentation) in robustness to such patch attacks.
On ImageNet, shift consistency is more pronounced than other methods, especially when larger patches are erased.

\begin{table}
\centering
\large
\resizebox{\linewidth}{!}{
\begin{tabular}{@{}lcccccc@{}}
\toprule
Method & Clean $\downarrow$ & mCE $\downarrow$ & Noise & Blur & Weather & Digital \\
\midrule
VGG-19                    & 25.8          & 81.6          & 84.33 & 86.50 & 72.00 & 84.00 \\ 
VGG-19 +\textbf{TIPS}     & \textbf{25.1} & \textbf{81.1} & 84.17 & 86.13 & 71.40 & 83.53 \\ 
\midrule
ResNet-18                 & 30.2          & 84.7          & 88.67 & 87.75 & 79.25 & 83.25 \\ 
ResNet-18 +\textbf{TIPS}  & \textbf{28.7} & \textbf{83.9} & 88.27 & 87.20 & 78.40 & 83.00 \\ 
\bottomrule
\end{tabular}
}
\caption{Errors on clean (ImageNet) and corrupted (ImageNet-C) test sets.
Models are trained only on clean data. 
}
\label{tab:corrupt}
\end{table}

\smallskip\noindent\textbf{Natural Corruptions}
We evaluated TIPS under an \textit{out-of-distribution} setting, where models are trained on clean images, but tested on images with natural corruptions~\citep{hendrycks2018benchmarking} due to noise, blur, weather, or digital artifacts.
\Tabref{tab:corrupt} shows that with TIPS, both the clean error and the mean corruption error (mCE) for VGG-19 and ResNet-18 improves.

\section{Conclusion} \label{sec:conc}

Through a large-scale correlation analysis, we identify a strong inverse relationship between shift invariance of CNNs and the maximum sampling bias (MSB) of pooling operators.
We find that optimizing model weights to reduce MSB is a good strategy for improving shift invariance.
We introduce the Translation Invariant Polyphase Sampling (TIPS) pooling layer and regularization to promote low MSB, resulting in state-of-the-art results for shift invariance on several image recognition benchmarks, outperforming data augmentation and contrastive learning strategies. 
Our analysis reveals additional benefits of TIPS such as improved robustness to adversarial attacks and corruptions.
Our work lays the foundation for further investigations on factors causing sensitivity to shifts of input signals in CNNs.

\medskip\noindent\textbf{Acknowledgments.}
The authors acknowledge support from UMBC ORCA, UMBC HPCF GPU cluster, and Kowshik Thopalli for valuable feedback.
The views and opinions of the authors expressed herein do not necessarily state or reflect those of the funding
agencies and employers.

{\small
\bibliographystyle{ieee_fullname}
\bibliography{egbib}
}

\clearpage

\twocolumn[
\centering
\Large
\textbf{Improving Shift Invariance in Convolutional Neural Networks with Translation Invariant Polyphase Sampling: \textit{Supplementary Material}}
\bigskip
\bigskip
]

\section{Appendix}
In this appendix, we define standard and circular shifts of images with examples and distinguish between shift invariance, equivariance, and non-invariance. 
We also discuss how polyphase decomposition operation in TIPS compares to previous works on signal propagation within CNNs for visual recognition. 
We further discuss computational analysis and experimental setup for MSB - shift invariance correlation study, image classification benchmarks, semantic segmentation benchmarks.
Finally, we illustrate the computational overhead in TIPS and discuss how it compares to existing pooling operators.

\begin{figure}[h]
    \centering
    \includegraphics[width=\linewidth]{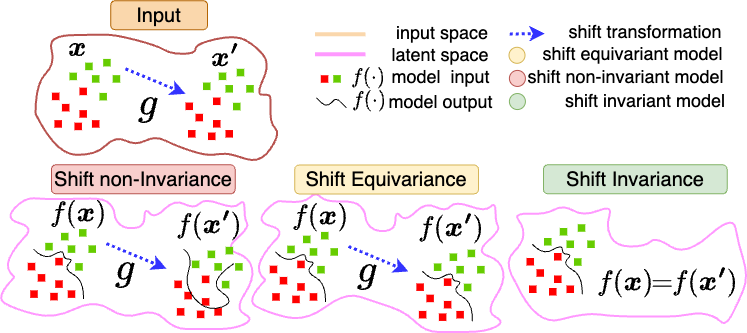}
    \caption{
    An illustration of shift equivariance, non-invariance, and invariance. 
    Invariant models map shifted, non-shifted inputs to identical outputs, while equivariant models mirror the input shift in outputs.
    }
    \label{fig:equivariance}
\end{figure}

\subsection{Shift Equivariance, Invariance, and non-Invariance}  \label{sec:shift_equi}
\Figref{fig:equivariance} depicts three scenarios where an input $x$ undergoes a transformation $g$ before being fed into a model $f$ to generate a prediction $\hat{y} = f(g(x)) = g^\prime (f(x))$: 
shift equivariance, shift non-invariance, and shift invariance.
If $g^\prime = g$, then $f$ is $g$-equivariant and if $g^\prime = I$ then $f$ is $g$-invariant.
Shift-invariance is desirable for image classification to ensure that categorical outputs are invariant to pixel shift, and shift-equivariance is desirable for semantic segmentation and object detection to ensure that pixel-shift in the image results in equivalent shift in corresponding segmentation masks and bounding boxes.

\begin{figure*}[h]
    \centering
    \includegraphics[width=0.8\textwidth]{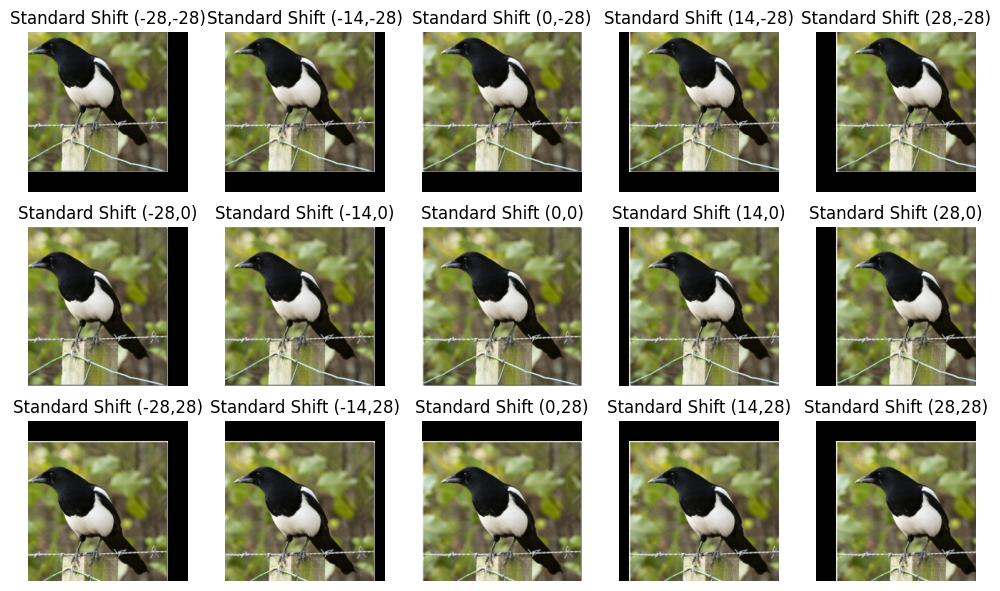} 
    \caption{
    Standard shift of an $224 \times 224$ image from ImageNet test set is shown with varying amount of shifts. Here, standard shift ($0,0$) denotes the original image with no shifts. It is also observed that, as the amount of standard shift increases, there occurs more information (pixel) loss.
    }
    \label{fig:std_shift_example}
\end{figure*}

\begin{figure*}[h]
    \centering
    \includegraphics[width=0.8\textwidth]{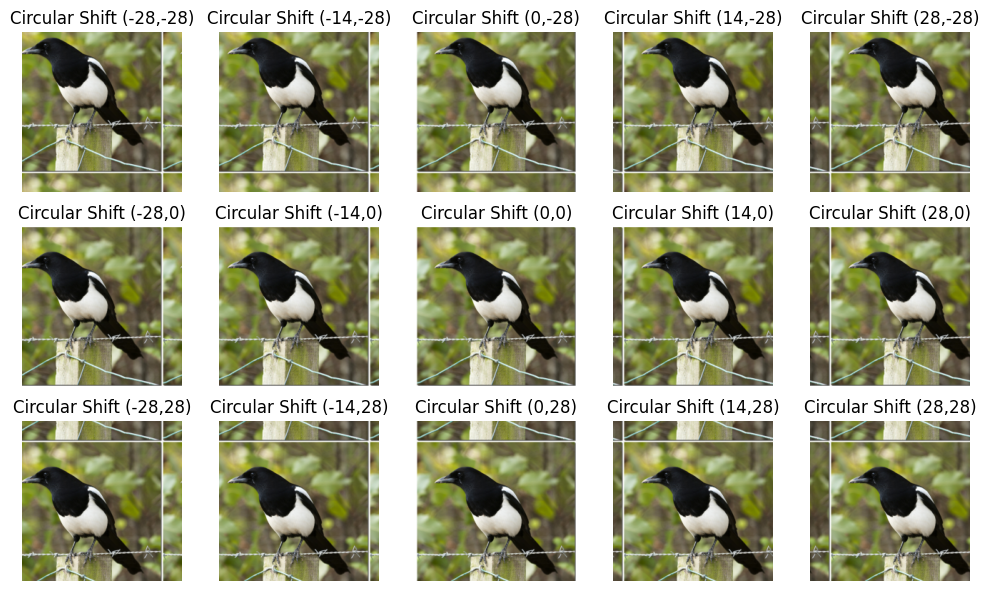} 
    \caption{
    Circular shift of an $224 \times 224$ image from ImageNet test set is shown with varying amount of shifts. Here, circular shift ($0,0$) denotes the original image with no shifts.
    }
    \label{fig:cir_shift_example}
\end{figure*}

\subsection{Standard and Circular Shifts of Images}  \label{sec:shift_types}
There are two types of pixel levels shifts that can be performed on images: standard shift and circular shift.
Given an image of height $h$ and width $w$, we can perform either type of shifts by an amount $(x,y)$ where $x \in \{0,..,h\}$, $y \in \{0,..,w\}$.
Standard shift is the process of shifting images to a $(x,y)$ direction which renders blank pixels at shifted positions.
Circular shift also shifts images in the $(x,y)$ direction, except the shifted pixels that move beyond the image boundary, are wrapped about the opposite ends of the image to fill in the empty pixels.
Therefore, circular shift is a lossless transformation while standard shift is not.
\Figref{fig:std_shift_example} and~\ref{fig:cir_shift_example} show examples of standard and circular shift (by varying amounts) applied to an image taken from ImageNet test set and  depict how standard shift renders blank pixels while circular shift do not.

\begin{table*}[t]
\resizebox{\linewidth}{!}{
\centering
\subfloat[Food-101]{
\resizebox{0.5\linewidth}{!}{
\Large
\begin{tabular}{@{}cllllll@{}}
\toprule
& & \multicolumn{1}{c}{\textbf{Unshifted}} & \multicolumn{2}{c}{\textbf{Standard Shift}} & \multicolumn{2}{c}{\textbf{Circular Shift}} \\
\cmidrule(lr){3-3} \cmidrule(lr){4-5} \cmidrule(lr){6-7}
 &\textbf{Method} & \textbf{Acc.} $\uparrow$ & \textbf{Consistency} $\uparrow$ & \textbf{Fidelity} $\uparrow$ & \textbf{Consistency} $\uparrow$ & \textbf{Fidelity} $\uparrow$ \\
\midrule
\multirow{8}{*}{\rotatebox[origin=c]{90}{CNN (ResNet-50)}} & MaxPool & 92.96{\footnotesize$\pm$0.08} & 82.13{\footnotesize$\pm$0.57} & 76.18{\footnotesize$\pm$0.07} & 83.61{\footnotesize$\pm$0.12} & 77.72{\footnotesize$\pm$0.05} \\
& APS & 94.68{\footnotesize$\pm$0.11} & 91.34{\footnotesize$\pm$0.04} & 86.48{\footnotesize$\pm$0.13} & \textbf{100.00{\footnotesize$\pm$0.00}} & 94.68{\footnotesize$\pm$0.11} \\
& LPS & 94.71{\footnotesize$\pm$0.02} & 92.41{\footnotesize$\pm$0.03} & 87.52{\footnotesize$\pm$0.11} & 99.48{\footnotesize$\pm$0.11} & 94.22{\footnotesize$\pm$0.05} \\
& \textbf{TIPS} & \textbf{95.63{\footnotesize$\pm$0.15}} & \textbf{95.02{\footnotesize$\pm$0.09}} & \textbf{90.87{\footnotesize$\pm$1.08}} & \textbf{100.00{\footnotesize$\pm$0.00}} & \textbf{95.63{\footnotesize$\pm$0.15}} \\
\cmidrule(lr){2-7}
& BlurPool (LPF-5) & 93.77{\footnotesize$\pm$0.03} & 88.18{\footnotesize$\pm$0.17} & 82.69{\footnotesize$\pm$1.08} & 93.49{\footnotesize$\pm$0.13} & 87.67{\footnotesize$\pm$0.03} \\
& APS (LPF-5) & 94.07{\footnotesize$\pm$0.13} & 92.51{\footnotesize$\pm$0.06} & 87.03{\footnotesize$\pm$0.20} & \textbf{100.00{\footnotesize$\pm$0.00}} & 94.07{\footnotesize$\pm$0.13} \\
& LPS (LPF-5) & 95.62{\footnotesize$\pm$0.07} & 94.10{\footnotesize$\pm$0.07} & 89.99{\footnotesize$\pm$0.19} & \textbf{100.00{\footnotesize$\pm$0.00}} & 95.62{\footnotesize$\pm$0.07} \\
& \textbf{TIPS (LPF-5)} & \textbf{96.42{\footnotesize$\pm$0.16}} & \textbf{95.50{\footnotesize$\pm$0.13}} & \textbf{92.08{\footnotesize$\pm$0.19}} & \textbf{100.00{\footnotesize$\pm$0.00}} & \textbf{96.42{\footnotesize$\pm$0.16}} \\
\midrule
\multirow{2}{*}{\rotatebox[origin=c]{90}{ViT}} & ViT-B/16 (I21k) & 96.88{\footnotesize$\pm$0.13} & 81.45{\footnotesize$\pm$0.04} & 78.91{\footnotesize$\pm$0.15} & \textbf{78.39{\footnotesize$\pm$0.12}} & 75.94{\footnotesize$\pm$0.12} \\
& ViT-L/16 (I21k) & 97.00{\footnotesize$\pm$0.03} & 81.84{\footnotesize$\pm$0.11} & 79.38{\footnotesize$\pm$0.08} & 78.06{\footnotesize$\pm$0.18} & 75.72{\footnotesize$\pm$0.17} \\
& Swin-B (I21k) & \textbf{97.49{\footnotesize$\pm$0.05}} & \textbf{82.85{\footnotesize$\pm$0.14}} & \textbf{80.77{\footnotesize$\pm$0.09}} & 78.05{\footnotesize$\pm$0.02} & \textbf{76.10{\footnotesize$\pm$0.08}} \\
\bottomrule
\end{tabular}
}
}%
\subfloat[Oxford-102]{
\resizebox{0.5\linewidth}{!}{
\Large
\begin{tabular}{@{}cllllll@{}}
\toprule
& & \multicolumn{1}{c}{\textbf{Unshifted}} & \multicolumn{2}{c}{\textbf{Standard Shift}} & \multicolumn{2}{c}{\textbf{Circular Shift}} \\
\cmidrule(lr){3-3} \cmidrule(lr){4-5} \cmidrule(lr){6-7}
 &\textbf{Method} & \textbf{Acc.} $\uparrow$ & \textbf{Consistency} $\uparrow$ & \textbf{Fidelity} $\uparrow$ & \textbf{Consistency} $\uparrow$ & \textbf{Fidelity} $\uparrow$ \\
\midrule
\multirow{8}{*}{\rotatebox[origin=c]{90}{CNN (ResNet-50)}} & MaxPool & 93.48{\footnotesize$\pm$0.15} & 85.63{\footnotesize$\pm$0.11} & 80.05{\footnotesize$\pm$0.17} & 89.38{\footnotesize$\pm$0.17} & 83.55{\footnotesize$\pm$0.12} \\
& APS & 94.68{\footnotesize$\pm$0.03} & 92.47{\footnotesize$\pm$0.05} & 87.55{\footnotesize$\pm$1.09} & \textbf{100.00{\footnotesize$\pm$0.00}} & 94.68{\footnotesize$\pm$0.03} \\
& LPS & 95.31{\footnotesize$\pm$0.08} & 93.63{\footnotesize$\pm$0.17} & 89.24{\footnotesize$\pm$0.11} & \textbf{100.00{\footnotesize$\pm$0.00}} & 95.31{\footnotesize$\pm$0.08} \\
& \textbf{TIPS} & \textbf{97.18{\footnotesize$\pm$0.06}} & \textbf{95.78{\footnotesize$\pm$0.03}} & \textbf{93.08{\footnotesize$\pm$0.16}} & \textbf{100.00{\footnotesize$\pm$0.00}} & \textbf{97.18{\footnotesize$\pm$0.06}} \\
\cmidrule(lr){2-7}
& BlurPool (LPF-5) & 92.71{\footnotesize$\pm$0.08} & 90.32{\footnotesize$\pm$0.13} & 83.74{\footnotesize$\pm$0.05} & 94.07{\footnotesize$\pm$0.13} & 87.21{\footnotesize$\pm$0.08} \\
& APS (LPF-5) & 94.71{\footnotesize$\pm$0.11} & 93.00{\footnotesize$\pm$0.08} & 88.09{\footnotesize$\pm$0.14} & \textbf{100.00{\footnotesize$\pm$0.00}} & 94.71{\footnotesize$\pm$0.11} \\
& LPS (LPF-5) & 96.28{\footnotesize$\pm$0.05} & 94.33{\footnotesize$\pm$0.06} & 90.82{\footnotesize$\pm$0.09} & \textbf{100.00{\footnotesize$\pm$0.00}} & 96.28{\footnotesize$\pm$0.05} \\ 
& \textbf{TIPS (LPF-5)} & \textbf{97.62{\footnotesize$\pm$0.11}} & \textbf{96.51{\footnotesize$\pm$0.14}} & \textbf{94.21{\footnotesize$\pm$0.14}} & \textbf{100.00{\footnotesize$\pm$0.00}} & \textbf{97.62{\footnotesize$\pm$0.11}} \\
\midrule
\multirow{2}{*}{\rotatebox[origin=c]{90}{ViT}} & ViT-B/16 (I21k) & 99.33{\footnotesize$\pm$0.05} & \textbf{88.47{\footnotesize$\pm$0.04}} & \textbf{87.88{\footnotesize$\pm$0.08}} & 82.24{\footnotesize$\pm$0.03} & 81.69{\footnotesize$\pm$0.06} \\
& ViT-L/16 (I21k) & 99.59{\footnotesize$\pm$0.03} & 87.25{\footnotesize$\pm$0.09} & 86.89{\footnotesize$\pm$0.18} & 82.39{\footnotesize$\pm$0.13} & 82.05{\footnotesize$\pm$0.03} \\
& Swin-B (I21k) & \textbf{99.68{\footnotesize$\pm$0.02}} & 87.06{\footnotesize$\pm$0.16} & 80.16{\footnotesize$\pm$0.07} & \textbf{83.57{\footnotesize$\pm$0.11}} & \textbf{83.30{\footnotesize$\pm$0.05}} \\
\bottomrule
\end{tabular}
}
}
}
\caption{
Image classification performance on Food-101 and Oxford-102 datasets averaged over five trials.
}
\label{tab:cls_food}
\end{table*}
\begin{table*}[t]
\centering
\large
\resizebox{\linewidth}{!}{
\begin{tabular}{lcllllllllll}
\toprule
\multicolumn{1}{c}{} & \multicolumn{1}{c}{} & \multicolumn{5}{c}{\textbf{Kvasir - U-Net}} & \multicolumn{5}{c}{\textbf{CVC-ClinicDB - U-Net}} \\
\cmidrule(lr){3-7} \cmidrule(lr){8-12}
\multicolumn{2}{c}{} & \multicolumn{1}{c}{\textbf{Unshifted}} & \multicolumn{2}{c}{\textbf{Standard Shift}} & \multicolumn{2}{c}{\textbf{Circular Shift}} & \multicolumn{1}{c}{\textbf{Unshifted}} & \multicolumn{2}{c}{\textbf{Standard Shift}} & \multicolumn{2}{c}{\textbf{Circular Shift}} \\
\cmidrule(lr){3-3} \cmidrule(lr){4-5} \cmidrule(lr){6-7} \cmidrule(lr){8-8} \cmidrule(lr){9-10} \cmidrule(lr){11-12}
\textbf{Method} & \textbf{Anti-Alias} & \textbf{mIOU} $\uparrow$ & \textbf{Consistency} $\uparrow$ & \textbf{Fidelity} $\uparrow$ & \textbf{Consistency} $\uparrow$ & \textbf{Fidelity} $\uparrow$ & \textbf{mIOU} $\uparrow$ & \textbf{Consistency} $\uparrow$ & \textbf{Fidelity} $\uparrow$ & \textbf{Consistency} $\uparrow$ & \textbf{Fidelity} $\uparrow$ \\
\midrule
MaxPool       &   -   & 75.60 & 92.84 & 70.19 & 97.91 & 74.02 & 73.81 & 90.24 & 66.61 & 95.50 & 70.50 \\
Blurpool      & LPF-3 & 78.39 & 94.63 & 74.18 & 98.30 & 77.06 & 76.32 & 93.87 & 71.64 & 96.36 & 73.54 \\
DDAC          & LPF-3 & 79.24 & 95.17 & 75.41 & 98.49 & 78.04 & 77.89 & 92.17 & 71.80 & 97.73 & 76.12 \\
APS           & LPF-3 & 81.97 & 96.32 & 78.95 &\textbf{100.00 }& 81.97 & 79.31 & 95.63 & 75.84 &\textbf{100.00} & 79.31 \\
LPS           & LPF-3 & 82.38 & 97.86 & 80.62 &\textbf{100.00 }& 82.38 & 78.59 & 96.21 & 75.61 &\textbf{100.00} & 78.59 \\
\textbf{TIPS} & LPF-3 & \textbf{86.10} & \textbf{98.09} & \textbf{84.46} &\textbf{100.00} & \textbf{86.10} & \textbf{80.05} & \textbf{97.89} & \textbf{78.36} &\textbf{100.00} & \textbf{80.05} \\
\bottomrule
\end{tabular}
}
\caption{Semantic segmentation performance on Kvasir and CVC-ClinicDB datasets.}
\label{tab:seg2}
\end{table*}

\begin{figure*}[!ht]
    \centering
    \includegraphics[width=\linewidth,trim={0 0em 0 0},clip]{images/seg_result.png}
    \caption{
    Qualitative comparison of segmentation masks predicted on original and shifted images. 
    Images from Cityscapes, Pascal VOC are standard-shifted by (43,-17), (-38,0) respectively.
    Regions where TIPS achieve improvements (i.e. consistent segmentation quality) under linear shifts are highlighted with circles.
    }
    \label{fig:seg_supp}
\end{figure*}

\subsection{Comparison of Polyphase Decomposition operation in the TIPS layer with previous work in signal propagation within CNNs for image recognition}  \label{sec:poly_decomp_and_prev_work}
Within TIPS layers, we use polyphase decomposition which is comparable to dilated convolution~\cite{yu2015multi} and dilated attention~\cite{jiao2023dilateformer} where the stride and dilation rates are identical (Fig 2 in manuscript). 
Usage of strided convolution in the above convolution operations can also be used for spatial downsampling, however strided convolutions are still shift invariant~\cite{zhang2019making}.
The slicing operation in polyphase decomposition is also identical to that of parallel grid pooling~\cite{takeki2018parallel}, focus layer in YOLOv5~\cite{hussain2024yolov1}. 
However, we learn to sample from these polyphase decompositions in the channel dimension while~\cite{takeki2018parallel, hussain2024yolov1} stack these decompositions in the channel space and then uses group convolution to downsample across the channel dimension.

\begin{table*}[ht]
\centering
\small
\begin{tabular}{@{}llllll@{}}
\toprule
\multicolumn{3}{c}{\textbf{Image Classification Experiments}} & \multicolumn{3}{c}{\textbf{Semantic Segmentation Experiments}} \\
\cmidrule(lr){1-3} \cmidrule(lr){4-6} 
\textbf{Model} & \textbf{\# Layers} & \textbf{Dataset}& \textbf{Model} & \textbf{\# Layers} & \textbf{Dataset}  \\
\midrule
MobileNet  & $\{2,3,4,5\}$ & CIFAR-10   & DeepLabV3+ (ResNet-18)  & $\{2,3,4,5\}$ & PASCAL VOC 2012 \\
ResNet-18  & $\{2,3,4,5\}$ & CIFAR-100  & DeepLabV3+ (ResNet-101) & $\{3,4,5.6\}$ & Cityscapes      \\
ResNet-34  & $\{2,3,4,5\}$ & Food-101   & U-Net (ResNet-18)       & $\{2,3,4,5\}$ & Kvasir          \\
ResNet-101 & $\{2,3,4,5\}$ & Oxford-102 & U-Net (ResNet-34)       & $\{2,3,4,5\}$ & CVC-ClinicDB    \\
\bottomrule
\end{tabular}
\caption{
List of CNN architectures and datasets, tested on each pooling method for correlation analysis between MSB and Shift Invariance.
}
\label{tab:correlation_setup}
\end{table*}
\begin{table*}[ht]
\centering
\resizebox{\textwidth}{!}{
\begin{tabular}{lllllllllllllllll}
\toprule
\multicolumn{1}{c}{} & \multicolumn{4}{c}{\textbf{CIFAR-10}} & \multicolumn{4}{c}{\textbf{CIFAR-100}} & \multicolumn{4}{c}{\textbf{Food-101}} & \multicolumn{4}{c}{\textbf{Oxford-102}} \\
\cmidrule(lr){2-5} \cmidrule(lr){6-9} \cmidrule(lr){10-13} \cmidrule(lr){14-17}
\textbf{Model}       & \textbf{$h\times w$} & \textbf{$b$} & \textbf{$s$}& \textbf{$N$}  & \textbf{$h\times w$} & \textbf{$b$} & \textbf{$s$}& \textbf{$N$}  & \textbf{$h\times w$} & \textbf{$b$} & \textbf{$s$}& \textbf{$N$}  & \textbf{$h\times w$} & \textbf{$b$} & \textbf{$s$}& \textbf{$N$}  \\
\midrule
MobileNet           & 32$\times$32 & 64 &  60 & 220 & 32$\times$32 & 64 &  60 & 220 & 200$\times$200 & 128&  60 & 220 & 200$\times$200 & 128&  60 & 220 \\
ResNet-18           & 32$\times$32 & 64 &  50 & 250 & 32$\times$32 & 64 &  50 & 250 & 224$\times$224 & 64 &  50 & 250 & 224$\times$224 & 64 &  50 & 250 \\
ResNet-34           & 32$\times$32 & 64 &  50 & 250 & 32$\times$32 & 64 &  50 & 250 & 224$\times$224 & 64 &  50 & 250 & 224$\times$224 & 64 &  50 & 250 \\
ResNet-101          & 32$\times$32 & 64 & 180 & 480 & 32$\times$32 & 64 & 180 & 480 & 224$\times$224 & 64 & 180 & 480 & 224$\times$224 & 64 & 180 & 480 \\
\bottomrule
\end{tabular}
}
\caption{
Image size ($h\times w$), batch size ($b$), step size($s$) for updating learning rate, and number of epochs ($N$) reported for each CNN model and image classification dataset combination for the MSB -- Shift Invariance correlation analysis experiment.
}
\label{tab:corr_cls_hyper}
\end{table*}
\begin{table*}[ht]
\centering
\resizebox{\textwidth}{!}{
\begin{tabular}{lllllllllllllllll}
\toprule
\multicolumn{1}{c}{} & \multicolumn{4}{c}{\textbf{Pascal VOC 2012}} & \multicolumn{4}{c}{\textbf{Cityscapes}} & \multicolumn{4}{c}{\textbf{Kvasir}} & \multicolumn{4}{c}{\textbf{CVC-ClinicDB}} \\
\cmidrule(lr){2-5} \cmidrule(lr){6-9} \cmidrule(lr){10-13} \cmidrule(lr){14-17}
\textbf{Model}       & \textbf{$h\times w$} & \textbf{$b$} & \textbf{$s$}& \textbf{$N$}  & \textbf{$h\times w$} & \textbf{$b$} & \textbf{$s$}& \textbf{$N$}  & \textbf{$h\times w$} & \textbf{$b$} & \textbf{$s$}& \textbf{$N$}  & \textbf{$h\times w$} & \textbf{$b$} & \textbf{$s$}& \textbf{$N$}  \\
\midrule
DeepLabV3+ (ResNet-18)     & 200$\times$300 & 12 &  120 & 450 & 200$\times$200 & 12 &  120 & 450 & 200$\times$200 & 12&  60 & 450 & 200$\times$300 &  8&  45 & 450 \\
DeepLabV3+ (ResNet-101)    & 200$\times$300 &  8 &  120 & 380 & 200$\times$200 & 12 &  120 & 380 & 200$\times$200 & 12 &  60 & 380 & 200$\times$300 &  8 &  45 & 380 \\
U-Net (ResNet-18)          & 200$\times$300 & 12 &  120 & 180 & 200$\times$200 & 16 &  120 & 180 & 200$\times$200 & 16 &  60 & 180 & 200$\times$300 & 12 &  45 & 180 \\
U-Net (ResNet-34)          & 200$\times$300 & 12 &  120 & 150 & 200$\times$200 & 12 &  120 & 150 & 200$\times$200 & 12 &  60 & 150 & 200$\times$300 &  8 & 45 & 150 \\
\bottomrule
\end{tabular}
}
\caption{
Image size ($h\times w$), batch size ($b$), step size($s$) for updating learning rate, and number of epochs ($N$) reported for each CNN model and semantic segmentation dataset combination for the MSB -- Shift Invariance correlation analysis experiment.
}
\label{tab:corr_seg_hyper}
\end{table*}

\subsection{Comparison of TIPS with existing Polyphase Sampling Pooling} \label{sec:tips_poly}

While TIPS, APS, and LPS use polyphase decomposition for spatial downsampling, they differ in how the pooled features are sampled from the decomposed polyphase components.  

\begin{itemize}[left=0pt, nosep]
    \item \textit{APS}  simply samples the polyphase component that contains the maximum energy using $\ell_{p}$ norm.
    \item \textit{LPS} learns to sample from these polyphase components using a shared convolution layer and gumble softmax.
    \item \textit{In TIPS}, the shared small convolution layer differs in design (Figure 2 in manuscript) from LPS. 
    In TIPS layers, we use convolution kernels, Global Average Pooling (GAP) layer and softmax activation that learns mixing coefficients (eqn 2 in manuscript) to sample polyphase components avoiding the sensitivity to gumble softmax temperature. 
\end{itemize}

\subsection{More Results on Image Classification and Semantic Segmentation} \label{sec:more_results}
\Tabref{tab:cls_food} contains results from image classification experiments on Food-101 and Oxford-102 datasets. 
\Tabref{tab:seg2} contains quantitative results from semantic segmentation experiments on Kvasir and CVC-ClinicDB datasets while \Figref{fig:seg_supp} contains qualitative results from semantic segmentation experiments on Pascal VOC and Cityscapes datasets.

\subsection{Experimental Setup for MSB - Shift Invariance Correlation Study}  \label{sec:corr_exp_details}
\Tabref{tab:correlation_setup} shows the list of CNN architectures (including Mobile Net~\cite{howard2017mobilenets}), datasets and pooling methods that we use to obtain a total of 576 configurations for the \textit{MSB-shift invariance} correlation study. 
In our study, we train each combination of architecture and dataset on 9 pooling methods: Global Average Pooling before classification with no spatial downsampling of convolution features, TIPS ($\epsilon=0.4, \alpha=0.35$), LPS ($\tau=0.01$), APS ($p=2$), APS ($p \to \infty$), LPS ($\tau \to \infty$), BlurPool (LPF-5), Average Pool ($2 \times 2$), and MaxPool ($2 \times 2$). 
Furthermore, in each of the aforementioned settings, we use different number of pooling layers as shown in \Tabref{tab:correlation_setup}.
While training with Global Average Pooling, we use 4 different kernel sizes ($2 \times 2, 3 \times 3, 4 \times 4, 5 \times 5$) in the first convolution layer with \textit{same padding} to create 4 variants since varying the number of pooling layers is not possible in this setting barring that we downsample only once (downsampling the very last convolution features with Global Average Pooling before classification/segmentation layer). 
Furthermore, to consider a wide variety of CNN design strategies in our correlation study, we use ResNets which has architectural choices such as residual / skip connections with varying depth and MobileNet which contains group (depthwise and separable) convolutions. 
Additionally, to make our correlation study more robust we consider more diverse configurations such as number of pooling layers, different datasets with varying magnitude of image resolution, number of classes etc. 
Moreover, we train and test all of these 576 configurations which is computationally expensive while using other CNN architectures such as VGG-16~\cite{simonyan2015very}, ConvNext~\cite{liu2022convnet}.

\begin{table*}[ht]
\centering
\resizebox{\linewidth}{!}{
\subfloat[Image Classification]{
\begin{tabular}{llccc}
\toprule
\textbf{Architecture}   & \textbf{Pooling} & \textbf{CUDA Time $\downarrow$} & \textbf{CUDA Memory $\downarrow$} & \textbf{GFLOPs $\downarrow$} \\
\midrule
MobileNet                           & MaxPool &      0.635  &      58.122  &        2.270  \\
                                    & TIPS    &      1.045  &     101.214  &        3.005  \\
                                    & GAP     &     66.609  &    6390.284  &      639.259  \\
\midrule
ResNet-18                           & MaxPool &      1.135  &      21.860  &        4.017  \\
                                    & TIPS    &      3.525  &     292.844  &        41.937 \\
                                    & GAP     &     72.460  &    1957.691  &      1124.032 \\
\midrule
ResNet-34                           & MaxPool &      1.954  &      31.904  &         8.128 \\
                                    & TIPS    &      5.623  &     334.754  &        71.532 \\
                                    & GAP     &     141.075 &    3451.912  &      2250.287 \\
\midrule
ResNet-101                          & MaxPool &      4.921  &     131.035  &        31.197 \\
                                    & TIPS    &     12.204  &     816.791  &       146.596 \\
                                    & GAP     &    534.514  &   21144.011  &      8508.809 \\
\bottomrule
\end{tabular}
}

\subfloat[Semantic Segmentation]{
\begin{tabular}{llccc}
\toprule
\textbf{Architecture}   & \textbf{Pooling} & \textbf{CUDA Time $\downarrow$} & \textbf{CUDA Memory $\downarrow$} & \textbf{GFLOPs $\downarrow$} \\
\midrule
DeepLabV3+(ResNet-18)               & MaxPool &      1.33   &      25.216  &         9.570 \\
                                    & TIPS    &      3.728  &     380.146  &        91.146 \\
                                    & GAP     &    121.029  &    2453.834  &      1926.592 \\
\midrule
DeepLabV3+(ResNet-101)              & MaxPool &      7.52   &     144.737  &        51.447 \\
                                    & TIPS    &     18.274  &     911.845  &       246.947 \\
                                    & GAP     &    741.568  &   21671.086  &     11521.953 \\
\midrule
U-Net(ResNet-18)                    & MaxPool &      2.754  &      78.574  &        23.567 \\
                                    & TIPS    &      8.675  &    1113.227  &       235.797 \\
                                    & GAP     &    143.402  &    3137.765  &      2700.095 \\
\midrule
U-Net(ResNet-34)                    & MaxPool &      3.179  &      88.707  &        27.678 \\
                                    & TIPS    &      9.045  &     957.965  &       246.352 \\
                                    & GAP     &    202.312  &    4631.986  &      3826.350 \\
\bottomrule
\end{tabular}
}
}
\caption{
GPU resources (CUDA time, memory, GFLOPs) allocated to convolution operations in CNNs while using different pooling operators for various CNN architectures.
We observe that, performing Global Average Pooling (GAP) on the final convolution feature with no prior downsampling drastically increases GPU resources in comparison to baseline MaxPool.
TIPS require additional convolution layers (Figure 2, manuscript), since it is a learnable pooling operator.
Compared to MaxPool, the overhead in GPU resources with TIPS is remarkably smaller than it is for Global Average Pooling.
}
\label{tab:conv_mem_stat}
\end{table*}

In \Tabref{tab:corr_cls_hyper}, \Tabref{tab:corr_seg_hyper} we include training details such as image size, batch size, step size, number of training epochs for all model - dataset combinations used in the MSB - shift invariance correlation framework for both image classification and semantic segmentation.
As discussed in Section 4 (manuscript), using Global Average Pooling with no spatial downsampling of the convolution features leads to increased computation with larger spatial features.
In \Tabref{tab:conv_mem_stat}, we summarize a detailed analysis on how Global Average Pooling increases computational complexity in comparison to baseline MaxPool.
The reported CUDA time is in \textit{nanoseconds (ns)}, CUDA memory is in \textit{Mega Bytes (MB)}, GFLOPs is \textit{billions} of floating point operations per second.
In (Figure 4, manuscript), we observe that Global Average Pooling improves shift invariance and reduces MSB, and \Tabref{tab:conv_mem_stat} reveals that this performance gain comes at a significantly higher computational cost which is impractical.
However, with TIPS we achieve comparable shift invariance and MSB by introducing marginal computational complexity in comparison to Global Average Pooling.

\begin{table*}[ht]
\centering
\Large
\resizebox{\textwidth}{!}{
\begin{tabular}{lllllllll}
\toprule
\multicolumn{1}{c}{} & \multicolumn{8}{c}{\textbf{Image Classification Experiments}} \\
\cmidrule(lr){2-9} 
\textbf{Dataset} & \textbf{Model} & \textbf{Batch Size} & \textbf{Step Size} & \textbf{Epochs} & \textbf{Image Size} & \textbf{\# Classes} & \textbf{\# Training Samples} & \textbf{\# Validation Samples}  \\
\midrule
CIFAR-10       & ResNet-18  & 64 & 50  & 250 & 32$\times$32   & 10   & 50,000    & 10,000    \\
Food-101       & ResNet-50  & 64 & 25  & 80  & 224$\times$224 & 101  & 75,750    & 25,250    \\
Oxford-102     & ResNet-50  & 64 & 20  & 70  & 224$\times$224 & 102  & 2,060     & 6,129     \\
Tiny ImageNet  & ResNet-101 & 64 & 180 & 480 & 64$\times$64   & 200  & 100,000   & 10,000    \\
ImageNet       & ResNet-101 & 64 & 30  & 90  & 224$\times$224 & 1000 & 1,281,167 & 50,000    \\
\bottomrule
\end{tabular}
}
\caption{
Training details, dataset statistics for all five datasets in our image classification experiments.
Training details include batch size, step size for updating learning rate, number of training epochs, image size and dataset statistics include number of classes, training samples, validation samples.
}
\label{tab:exp_det_img_cls}
\end{table*}
\begin{table*}[ht]
\centering
\resizebox{\textwidth}{!}{
\begin{tabular}{lllllllll}
\toprule
\multicolumn{1}{c}{} & \multicolumn{8}{c}{\textbf{Semantic Segmentation Experiments}} \\
\cmidrule(lr){2-9} 
\textbf{Dataset} & \textbf{Model} & \textbf{Batch Size} & \textbf{Step Size} & \textbf{Epochs} & \textbf{Image Size} & \textbf{\# Classes} & \textbf{\# Training Samples} & \textbf{\# Validation Samples}  \\
\midrule
PASCAL VOC 2012  & DeepLabV3+(ResNet-18)  & 12 & 120 & 450 & 200$\times$300 & 20  & 1,464 & 1,456   \\
Cityscapes       & DeepLabV3+(ResNet-101) & 12 & 120 & 380 & 200$\times$200 & 19  & 2,975 & 500     \\
Kvasir           & UNet(ResNet-18)        & 12 & 60  & 180 & 200$\times$200 & 2   & 850   & 150     \\
CVC-ClinicDB     & UNet(ResNet-34)        & 8  & 45  & 150 & 200$\times$300 & 2   & 521   & 91      \\
\bottomrule
\end{tabular}
}
\caption{
Training details, dataset statistics for all four datasets in our semantic segmentation experiments.
Training details include batch size, step size for updating learning rate, number of training epochs, image size and dataset statistics include number of classes, training samples, validation samples.
}
\label{tab:exp_det_seg}
\end{table*}

\subsection{Experimental Setup for Image Classification, Object Detection, and Semantic Segmentation}  \label{sec:train_details}
We benchmark the performance of TIPS and prior work on five image classification datasets which are described in \Tabref{tab:exp_det_img_cls}.
We benchmark the performance of TIPS and prior work on four semantic segmentation datasets which are described in \Tabref{tab:exp_det_seg}.
\Tabref{tab:exp_det_img_cls},~\ref{tab:exp_det_seg} contains further training details on all the reported datasets such as batch size, step size, number of training epochs, image/crop size, number of classes and number of images in the dataset.

The values of $\epsilon = 0.4,~\alpha = 0.35$ were obtained using hyperparameter search on CIFAR-10. Note that we only run hyperparameter ($\epsilon$, $\alpha$) tuning on CIFAR-10 (a small dataset) and then use the same $\epsilon$, $\alpha$ for other image classification, object detection, and semantic segmentation benchmarks without any hyperparameter search on other tasks or datasets. 
We chose $\epsilon=0.4$ because introducing $\mathcal{L}_{undo}$ after 40\% of the training duration yields the best performance (see Fig 8 in manuscript).

\begin{table*}[ht]
\centering
\resizebox{\linewidth}{!}{
\subfloat[Image Classification]{
\begin{tabular}{lllll}
\toprule
\textbf{Method}         &     \textbf{ResNet-18} &     \textbf{ResNet-34} &       \textbf{ResNet-50} &     \textbf{ResNet-101}  \\
\midrule
BlurPool      &  0.00 &  0.00 & 0.00 & 0.00  \\
DDAC          &  7.92  &  10.53 & 9.27  & 4.30   \\
APS           &  0.00 &  0.00 & 0.00 & 0.00  \\
LPS           &  1.03  &  2.24  & 1.93  & 1.05   \\
\textbf{TIPS} &  5.51  &  4.56  & 2.17  & 3.19   \\
\bottomrule
\end{tabular}
}

\subfloat[Semantic Segmentation]{
\begin{tabular}{llll}
\toprule
\textbf{Method}   &  \textbf{DeepLabV3+(A)} &   \textbf{DeepLabV3+(B)} &         \textbf{UNet}   \\
\midrule
BlurPool       & 0.00 & 0.00 & 0.00 \\
DDAC           & 12.00 & 4.83  & 12.83 \\
APS            & 0.00 & 0.00 & 0.00 \\
LPS            & 4.40  & 3.25  & 4.79 \\
\textbf{TIPS}  & 7.24  & 4.04  & 5.76 \\
\bottomrule
\end{tabular}
}
}
\caption{
Percentage of additional parameters required by different pooling operators in comparison to MaxPool on each CNN architecture for classification and semantic segmentation. 
We observe that, while TIPS require more parameters than LPS, DDAC causes the maximum increase in trainable parameters \textit{w.r.t.} baseline MaxPool. 
}
\label{tab:params_relative}
\end{table*}

\begin{table*}[ht]
\centering
\resizebox{\linewidth}{!}{
\subfloat[Image Classification]{
\begin{tabular}{lllll}
\toprule
\textbf{Method}         &     \textbf{ResNet-18} &     \textbf{ResNet-34} &       \textbf{ResNet-50} &     \textbf{ResNet-101}  \\
\midrule
MaxPool       &  11.884 & 21.282 & 23.521 & 42.520 \\ 
BlurPool      &  11.884 & 21.282 & 23.521 & 42.520 \\   
DDAC          &  12.825 & 23.524 & 25.701 & 44.349 \\
APS           &  11.884 & 21.282 & 23.521 & 42.520 \\
LPS           &  12.006 & 21.759 & 23.975 & 42.966 \\
\textbf{TIPS} &  12.539 & 22.253 & 24.031 & 43.876 \\
\bottomrule
\end{tabular}
}

\subfloat[Semantic Segmentation]{
\begin{tabular}{llll}
\toprule
\textbf{Method}   &  \textbf{DeepLabV3+(A)} &   \textbf{DeepLabV3+(B)} &         \textbf{UNet}   \\
\midrule
MaxPool        & 20.131 & 58.630 & 7.762 \\
BlurPool       & 20.131 & 58.630 & 7.762 \\
DDAC           & 22.547 & 61.459 & 8.758 \\
APS            & 20.131 & 58.630 & 7.762 \\
LPS            & 21.017 & 60.536 & 8.134 \\
\textbf{TIPS}  & 21.589 & 60.999 & 8.209 \\
\bottomrule
\end{tabular}
}
}
\caption{
Number of trainable parameters in \textbf{Million} for various pooling methods reported for: ResNet-18, ResNet-34, ResNet-50, ResNet-101 backbones (image classification), DeepLabV3+ (A: ResNet-18, B: ResNet-101) and UNet (semantic segmentation). 
Number of trainable parameters are computed assuming an RGB input image of size 224 $\times$ 224.
}
\label{tab:params}
\end{table*}

\subsection{Computational Overhead in TIPS} \label{sec:param_study}
\Tabref{tab:params_relative} shows the percentage of additional parameters required to use TIPS on image classification and segmentation CNN models with different pooling methods and CNN architectures, for RGB images of size $224 \times 224 $ and a batch-size of 64.
TIPS introduces marginal computational overhead while still being computationally cheaper than existing pooling operators for shift invariance, i.e. DDAC.
Moreover, in \Tabref{tab:params} we show the number of trainable parameters with different pooling operators for all the image classification, semantic segmentation CNN models.
While TIPS requires higher number of trainable parameters than LPS, it is still much less than DDAC.

\begin{figure}[t]
    \centering
    \includegraphics[width=0.95\linewidth]{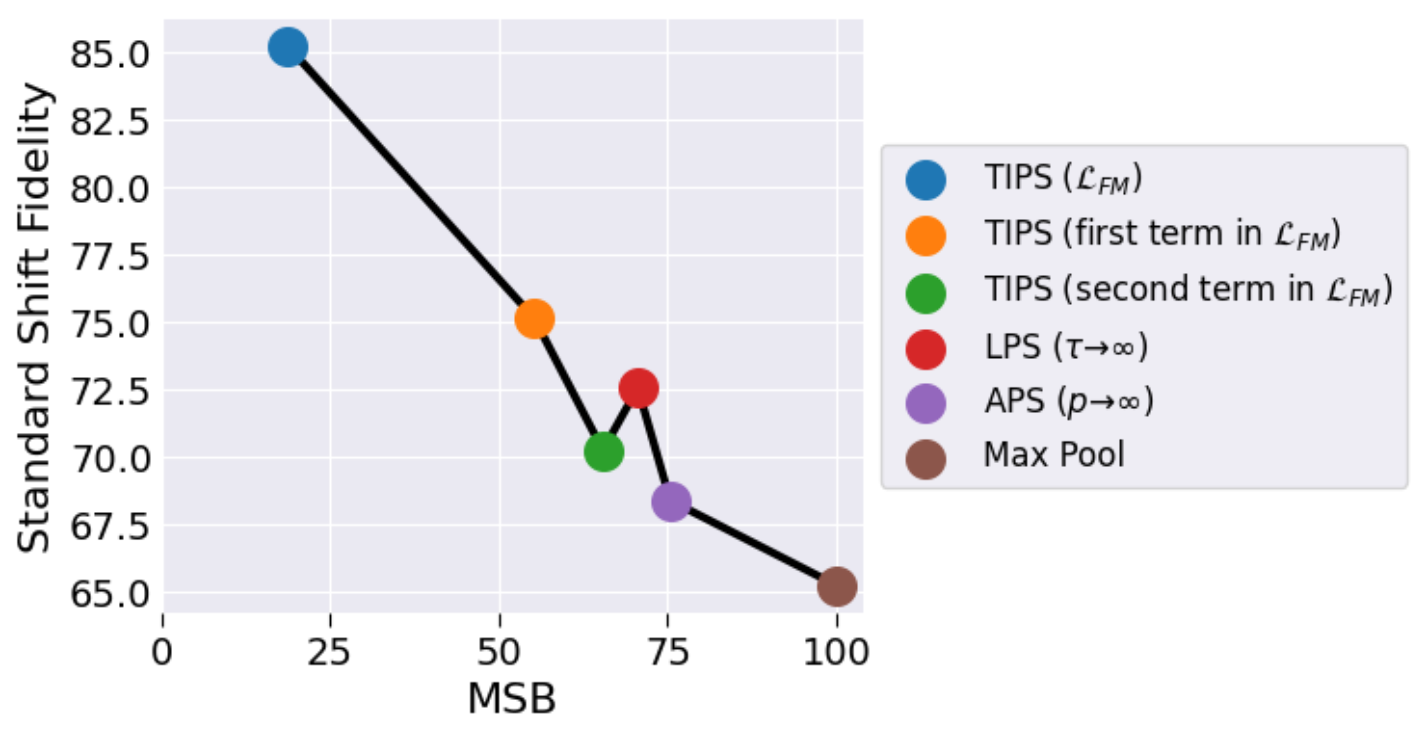}
    \caption{
    The effect of $\mathcal{L}_{FM}$ on TIPS is visualized by plotting standard shift fidelity versus MSB for models trained on Tiny ImageNet.
    Training TIPS with $\mathcal{L}_{FM}$ yields the maximum standard shift fidelity and minimum MSB.
    } 
    \label{fig:effect_LFM}
\end{figure}

\subsection{Effect of training on \texorpdfstring{$\mathcal{L}_{FM}$}{L_FM}} \label{sec:effect_Lfm}
In \Figref{fig:effect_LFM}, we train ResNet-101 on Tiny ImageNet with TIPS and $\mathcal{L}_{FM}$ and compare it with baselines LPS, APS and MaxPool in terms of standard fidelity and MSB.
To further inspect the effect of training TIPS with $\mathcal{L}_{FM}$, we train with three different setting of TIPS: (1) TIPS with $\mathcal{L}_{FM}$: to discourages both skewed and uniform $\tau$, (2) TIPS with only the first term in $\mathcal{L}_{FM}$: to discourages skewed $\tau$ only, and (3) TIPS with only second term in $\mathcal{L}_{FM}$: to discourages uniform $\tau$ only.
We observe that training TIPS with both terms from $\mathcal{L}_{FM}$ yields the maximum gain in shift fidelity and decreases MSB the most.
TIPS with $\mathcal{L}_{FM}$ also outperforms other pooling methods: LPS, APS and MaxPool in terms of standard shift fidelity and MSB.

\subsection{Studying Shift Invariance on CNN architectures beyond ResNets} \label{sec:other_CNNs}

Tables~\ref{tab:other_cnn_tab_1},~\ref{tab:other_cnn_tab_3}, contain results from DenseNet ~\cite{huang2017densely} and EfficientNet ~\cite{tan2019efficientnet} on CIFAR-10 with different pooling methods including TIPS. 
We use LPF-5 for antialiasing and hyperparameters used in DenseNet and EfficientNet respectively. 
We observe improved shift invariance with TIPS independent of CNN architecture.

\begin{table}[h]
\resizebox{\linewidth}{!}{
\centering
\large
\begin{tabular}{lccccc}
\toprule
\multicolumn{2}{c}{} & \multicolumn{2}{c}{\textbf{Standard Shift}} & \multicolumn{2}{c}{\textbf{Circular Shift}} \\ 
\cmidrule(lr){3-4} \cmidrule(lr){5-6}
\textbf{Method}  & \textbf{Accuracy} $\uparrow$ & \textbf{Consistency} $\uparrow$ & \textbf{Fidelity} $\uparrow$ & \textbf{Consistency} $\uparrow$ & \textbf{Fidelity} $\uparrow$ \\ 
\midrule
MaxPool                 & 96.37             & 90.02                & 86.72             & 92.41                & 89.05             \\ 
BlurPool                & 96.74             & 92.57                & 89.51             & 94.07                & 90.96             \\ 
APS                     & 97.27             & 93.31                & 90.82             & \textbf{100.00}               & 97.27             \\ 
LPS                     & 97.12             & 94.36                & 91.62             & \textbf{100.00}               & 97.12             \\ 
\textbf{TIPS}             & \textbf{97.43}    & \textbf{96.71}       & \textbf{94.19}    & \textbf{100.00}      & \textbf{97.43}    \\ 
\bottomrule
\end{tabular}
}
\caption{
Image classification performance on CIFAR-10 with DenseNet-BC (k=24)~\cite{huang2017densely}.
}
\label{tab:other_cnn_tab_1}
\end{table}
\begin{table}[h]
\resizebox{\linewidth}{!}{
\centering
\large
\begin{tabular}{lccccc}
\toprule
\multicolumn{2}{c}{} & \multicolumn{2}{c}{\textbf{Standard Shift}} & \multicolumn{2}{c}{\textbf{Circular Shift}} \\ 
\cmidrule(lr){3-4} \cmidrule(lr){5-6}
\textbf{Method}  & \textbf{Accuracy} $\uparrow$ & \textbf{Consistency} $\uparrow$ & \textbf{Fidelity} $\uparrow$ & \textbf{Consistency} $\uparrow$ & \textbf{Fidelity} $\uparrow$ \\ 
\midrule
MaxPool                 & 98.90             & 89.14                & 88.13             & 92.19                & 91.22             \\ 
BlurPool                & 98.90             & 91.06                & 90.07             & 92.37                & 91.39             \\ 
APS                     & 98.53             & 92.30                & 90.95             & 100.00               & 98.53             \\ 
LPS                     & 98.93             & 93.47                & 92.44             & 100.00               & 98.93             \\ 
\textbf{TIPS}             & \textbf{98.93}    & \textbf{93.67}       & \textbf{92.67}    & \textbf{100.00}      & \textbf{98.93}    \\ 
\bottomrule
\end{tabular}
}
\caption{
Image classification performance on CIFAR-10 with EfficientNet-B7~\cite{tan2019efficientnet}.
}
\label{tab:other_cnn_tab_3}
\end{table}

\subsection{Studying Effect of Normalization Layers on Shift Invariance} \label{sec:norm_effect}
Using normalization layers in CNNs positively impact visual recognition performance~\cite{ioffe2015batch, ba2016layer, wu2018group, nasirigerdehkernel}. 
However, the goal of this study is to carefully analyze (and isolate) the impact of pooling operators on shift invariance. 
While layer normalization is not the focus of this work, we have experimented on it's different alternatives and show results in Tables~\ref{tab:norm_layer_1},~\ref{tab:norm_layer_2}. 
Tables~\ref{tab:norm_layer_1} (batch size 32), ~\ref{tab:norm_layer_2} (batch size 256) contain results on CIFAR-10 with a ResNet-18 backbone with TIPS and MaxPool pooling with Batch Norm~\cite{ioffe2015batch}, Layer Norm~\cite{ba2016layer}, Group Norm~\cite{wu2018group}, and Kernel Norm~\cite{nasirigerdehkernel}. 
We observe that usage of normalization layers leads to mixed results – this points to normalization not being a major factor for shift invariance. 
However, Tables~\ref{tab:norm_layer_1},~\ref{tab:norm_layer_2} reveal that using TIPS instead of baseline MaxPool improves shift invariance regardless of layer normalization choice.

\begin{table}[h]
\resizebox{\linewidth}{!}{
\centering
\large
\begin{tabular}{lccccc}
\toprule
\multicolumn{2}{c}{} & \multicolumn{2}{c}{\textbf{Standard Shift}} & \multicolumn{2}{c}{\textbf{Circular Shift}} \\ 
\cmidrule(lr){3-4} \cmidrule(lr){5-6}
\textbf{Normalization}  & \textbf{Accuracy} $\uparrow$ & \textbf{Consistency} $\uparrow$ & \textbf{Fidelity} $\uparrow$ & \textbf{Consistency} $\uparrow$ & \textbf{Fidelity} $\uparrow$ \\ 
\midrule
Batch Norm~\cite{ioffe2015batch}  & 96.02/91.43 & \textbf{98.61}/87.43 & \textbf{94.69}/79.94 & \textbf{100.00}/90.18 & 96.02/82.45 \\ 
Layer Norm~\cite{ba2016layer}  & 93.43/92.25 & 97.34/\textbf{89.37} & 90.95/\textbf{89.77} & \textbf{100.00}/90.61 & 93.43/83.60 \\ 
Group Norm~\cite{wu2018group}  & 94.79/89.04 & 95.82/82.37 & 90.84/73.34 & \textbf{100.00}/\textbf{93.59} & 94.79/83.33 \\ 
Kernel Norm~\cite{nasirigerdehkernel} & \textbf{96.18}/\textbf{95.72} & 98.07/86.12 & 94.31/82.43 & \textbf{100.00}/90.81 & \textbf{96.18}/86.95 \\ 
\bottomrule
\end{tabular}
}
\caption{
Inspecting the influence of different layer normalization strategies on ResNet-18 for CIFAR-10 with different pooling operators. 
All results are reported as TIPS/MaxPool with batch size of 32. 
}
\label{tab:norm_layer_1}
\end{table}

\begin{table}[h]
\resizebox{\linewidth}{!}{
\centering
\large
\begin{tabular}{lccccc}
\toprule
\multicolumn{2}{c}{} & \multicolumn{2}{c}{\textbf{Standard Shift}} & \multicolumn{2}{c}{\textbf{Circular Shift}} \\ 
\cmidrule(lr){3-4} \cmidrule(lr){5-6}
\textbf{Normalization}  & \textbf{Accuracy} $\uparrow$ & \textbf{Consistency} $\uparrow$ & \textbf{Fidelity} $\uparrow$ & \textbf{Consistency} $\uparrow$ & \textbf{Fidelity} $\uparrow$ \\ 
\midrule
Batch Norm~\cite{ioffe2015batch}  & \textbf{94.71}/90.87 & 97.29/\textbf{88.29} & 92.15/80.21 & \textbf{100.00}/84.91 & \textbf{94.71}/76.89 \\ 
Layer Norm~\cite{ba2016layer}  & 94.67/91.19 & 96.43/82.13 & 91.29/74.91 & \textbf{100.00}/89.02 & 94.67/81.24 \\ 
Group Norm~\cite{wu2018group}  & 94.14/94.02 & 96.80/84.82 & 91.14/79.38 & \textbf{100.00}/92.30 & 94.14/86.69 \\ 
Kernel Norm~\cite{nasirigerdehkernel} & 94.58/\textbf{94.58} & \textbf{97.44}/86.36 & 92.16/81.69 & \textbf{100.00}/\textbf{93.47} & 94.58/\textbf{88.43} \\ 
\bottomrule
\end{tabular}
}
\caption{
Inspecting the influence of different layer normalization strategies on ResNet-18 for CIFAR-10 with different pooling operators. 
All results are reported as TIPS/MaxPool with batch size of 256. 
}
\label{tab:norm_layer_2}
\end{table}

\end{document}